\documentclass[11pt]{article}
\usepackage[utf8]{inputenc}
\pdfoutput=1
\usepackage{setspace}
\usepackage{palatino}
\usepackage{graphicx}
\usepackage{float}
\usepackage{titling} 
\usepackage{multirow}
\usepackage{lscape}
\usepackage{amsmath}
\usepackage{amssymb}
\usepackage[a4paper, total={6in, 9.5in}]{geometry}
\usepackage{array}
\usepackage[normalem]{ulem}
\fontfamily{ppl}\selectfont 
\usepackage{eqparbox}
\usepackage{arydshln}
\usepackage{mathrsfs}
\usepackage[american]{babel}
\usepackage{mathtools}

\usepackage{rotating}
\usepackage{amsthm}
\newtheorem{theorem}{Theorem}

\newtheorem{definition}{Definition}

\usepackage[ruled,vlined]{algorithm2e}

\usepackage{bm}
\usepackage{caption}
\usepackage{subcaption}

\usepackage[natbibapa]{apacite}
\PassOptionsToPackage{hyperindex,breaklinks}{hyperref}
\usepackage{hyperref}
\usepackage{xcolor}
\definecolor{darkblue}{rgb}{0, 0, 0.5}
\hypersetup{colorlinks=true,citecolor=darkblue, linkcolor=darkblue, urlcolor=darkblue}

\usepackage[textsize=small]{todonotes}

\title{Nonnegative matrix factorizations and related compositional models: Equivalence, identifiability, and an application on the grain-size analysis of sediments}

\author{Qianqian Qi\\
   {\small\raggedright Hangzhou Dianzi University, China}\\
   \href{mailto:q.qi@hdu.edu.cn}{\texttt{q.qi@hdu.edu.cn}} 
\and Peter G. M. van der Heijden\\
    {\small\raggedright Utrecht University, the Netherlands and
University of Southampton, UK}\\
\href{mailto:p.g.m.vanderheijden@uu.nl}{\texttt{p.g.m.vanderheijden@uu.nl}}
\and Maarten A. Prins\\
    {\small\raggedright Vrije Universiteit Amsterdam, the Netherlands}\\
\href{mailto:m.a.prins@vu.nl}{\texttt{m.a.prins@vu.nl}}
    }
    
\predate{}
\postdate{}
\date{}
\date{\vspace{-5ex}}

\begin{document}

{\setstretch{.8}
\maketitle
\begin{abstract}

Across fields such as machine learning, social science, and geology, considerable attention has been given to models that factorize a nonnegative matrix into the product of two or three matrices, subject to nonnegative or row-sum-to-1 constraints. Although these models are to a large extent similar or even equivalent, they are presented under different names, and their similarity is not well known. This paper highlights similarities among five models, latent budget analysis (LBA) and latent class analysis (LCA) from social science, end-member analysis (EMA) from geology, probabilistic latent semantic analysis (PLSA) and nonnegative matrix factorization (NMF) from machine learning. We focus on the identifiability of these models. We prove that the solution of LBA, EMA, LCA, PLSA is unique if and only if the solution of NMF is unique. Consequently, existing uniqueness theorems for NMF directly apply to LBA, EMA, LCA, PLSA, and vice versa. We also provide a brief review of algorithms for the estimation of these models. We illustrate NMF on a sedimentary grain-size distribution dataset from sedimentary geology, and end the paper with a discussion of closely related model: archetypal analysis.
\end{abstract}
}

\section{Introduction}\label{s: introduction}

In many settings in machine learning, geology, and social science, observed data are nonnegative and can be considered a mixture of multiple unobserved, or latent, sources \citep{SaberiMovahed2025Nonnegative, RENNY2026113384, VONOSMANSKI2026134139}. For these data we are interested in the latent sources and the mixing weights. To extract these latent structures, nonnegative matrix factorization (NMF) approximates the observed nonnegative matrix by a product of lower-rank nonnegative matrices. The seminal work of \citet{lee1999learning}, demonstrating that NMF can learn parts-based representations of objects, sparked much research in NMF. NMF is now recognized as a workhorse for data science \citep{gillis2020nonnegative, GUO2024102379, SaberiMovahed2025Nonnegative}.

In many applications, the observed data take the form of compositional data, where the elements in each row in a matrix add up to one \citep{Alenazi18082023}. To analyze such data, several models have been proposed independently across statistics and machine learning.

Latent budget analysis (LBA) is a statistical model known in the social sciences designed for the analysis of compositional data \citep{clogg1981latent, de1990latent, van1999identifiability, Alenazi18082023}. The name "latent budgets" stems from the original proposal to approximate so-called observed time budgets of (groups of) individuals by a number of latent, i.e. unknown time budgets. The term budget reflects that the elements of a budget add up to 1: comparable to a financial budget, one can spend an amount of time only once.

With a similar aim, 
in geology, the technique end-member analysis (EMA), having the same parametric form as LBA \citep{heijden1994end}, was proposed for deriving provenance, transport, and sedimentation processes from grain-size distribution data \citep{renner1993resolution,
weltje1997end,weltje2007genetically, RENNY2026113384}. End members refer to extreme source compositions. 

In machine learning, the technique probabilistic latent semantic analysis (PLSA), also called probabilistic latent semantic indexing (PLSI), whose asymmetric representation is also the same as LBA, was proposed for automated document indexing \citep{hofmann1999probabilistic, hofmann2001unsupervised, figuera2024revisiting}. PLSA is inspired by latent semantic analysis (LSA), another document indexing technique \citep{deerwester1990indexing}. Compared to LSA, PLSA defines a proper generative model for the data in terms of (conditional) probabilities.

LBA, EMA and asymmetric PLSA are models for compositional data, where the row elements add up to 1. There are two mathematically equivalent models for nonnegative matrices where all elements of the matrix add up to 1. These equivalences are well known: LBA is equivalent to latent class analysis (LCA) for two-way tables. \citep{lazarsfeld1968latent, clogg1981latent,
van1989latent}  and asymmetric PLSA is equivalent to symmetric PLSA \citep{hofmann1999probabilistic, hofmann2001unsupervised, figuera2024revisiting}.

Despite research on NMF, PLSA, EMA, LBA, and LCA for two-way tables, only few attempts have been made to establish the connections among them \citep{clogg1981latent, de1991reduced, heijden1994end, gaussier2005relation, ding2008equivalence, figuera2024revisiting}. These methods have largely developed independently across disciplines, limiting the exchange of theorems and algorithms. Establishing their connections can promote cross-disciplinary exchange.
Thus, one aim of this paper is to examine their relationship systematically. To the best of our knowledge, this is the first work linking LBA from social
science and EMA from geology to PLSA and NMF from computing science. 

One essential problem of NMF, PLSA, EMA, LBA, and LCA for two-way tables is model identifiability - under what conditions latent factors are unique  (up to permutation and scaling) \citep{goodman1987new, van1999identifiability, weltje1997end, gillis2020nonnegative}. The identifiability issue is tantamount to the question of whether or not these latent factors are the only interpretation of the data, or alternatives exist. In this paper, we show that the theoretical analyses of uniqueness for PLSA, EMA, LBA, and LCA for two-way tables are at different stages of development relative to those for NMF. We establish a theorem that allows uniqueness results to be directly transferred among these models.

In this paper we present an analysis of sediments. Sediments are the memory of dynamic processes acting at the earth surface \citep{van2018genetically, ZHANG2020106656}. Studying sediments and the sedimentary rock record allows earth scientists to reconstruct ancient environmental conditions, track modern sediment pathways, and predict future landscape responses to global change. Marine sediments are the building blocks of coastal landscapes. Waves, currents, and wind work together to move gravel, sand and mud across different temporal and spatial scales. They let coastlines grow, adapt, and protect the land from rising seas. Tracking sediment movement helps understanding the natural forces that create coastal landforms, and helps creating resilient shores that handle climate change safely.

The Slufter on the island of Texel is a uniquely dynamic, storm-driven tidal inlet system where unrestricted North Sea tides continually reshape the coastal landscape by erosion and deposition of sediments, offering an ideal natural laboratory to study morphodynamic resilience. Here we present a grain-size dataset which has been collected as part of a multi-year monitoring project initiated by the water board Hoogheemraadschap Hollands Noorderkwartier and government organization for forestry and the management of nature reserves Staatsbosbeheer. We will analyse this dataset using NMF from machine learning.

The paper is organized as follows.
Section~\ref{S: NMF, LBA, EMA, PLSA, and LCA} shows the relationship among LBA, EMA, asymmetric PLSA, and NMF, and the equivalent models LCA for two-way tables and symmetric PLSA. In Section~\ref{S: Identifiability issue}, we prove that the solution of NMF is unique if and only if the solutions of LBA, EMA, and PLSA are unique.  Section~\ref{S: Algorithms/Estimators} briefly introduces algorithms. Section~\ref{S: geologyexampleslufter} presents a geological application of NMF, the analysis of sediment transport in the Slufter tidal-inlet system. Section~\ref{S: Related models/methods} introduces related model: archetypal analysis. Finally, Section~\ref{S: Discussion and conclusion} concludes this paper.

\section{Relationships among NMF, PLSA, EMA, LBA, and LCA}\label{S: NMF, LBA, EMA, PLSA, and LCA}

In this section, we will show the relationships among NMF, PLSA, EMA, LBA, and LCA of two-way tables. Now we give a more technical introduction to NMF, PLSA, EMA, LBA, and LCA for two-way tables. Let $\bm{X}$ be an observed nonnegative matrix of size $I \times J$ with element $x_{ij}$ $(i =1,...,I; j = 1,...,J)$. Given a matrix $\bm{C}$, let $\bm{D}_{\bm{C}}$ be a diagonal matrix whose diagonal entries are the row sums of $\bm{C}$. For example, let $\bm{P} = \bm{D}_{\bm{X}}^{-1}\bm{X}$ with element $p_{ij} = x_{ij}/\sum_jx_{ij}$, where $\bm{D}_{\bm{X}}$ has elements $\sum_jx_{ij}$. $\bm{P}$ is also known as a matrix with compositional data, where the observed conditional row elements $p_{ij}$ for a row $i$ add up to 1. 

\subsection{NMF}

NMF approximates $\bm{X}$ by a lower rank matrix $\bm{\Phi}$ which is the product of two nonnegative matrices $\bm{M}$ and $\bm{H}$. I.e.,
\begin{equation}\label{E: nmfmod}
    \bm{\Phi} = \bm{M}\bm{H},
\end{equation}
\noindent where $\bm{M}$ and $\bm{H}$ have sizes $I\times K$ and $K\times J$ respectively and both have rank $K$ \citep{lee1999learning, gillis2020nonnegative}.
In (\ref{E: nmfmod}), $K$ is the so-called nonnegative rank of $\bm{\Phi}$.
The term ''nonnegative rank'' is used, instead of the term ''rank'', as $\bm{\Phi}$, $\bm{M}$, and $\bm{H}$ are all nonnegative. The nonnegative rank satisfies $K \le \text{min}(I,J)$, and if $K < \text{min}(I,J)$, then $\bm{\Phi}$ is a matrix of lower rank.  Due to the non-negativity constraints of $\bm{M}$ and $\bm{H}$, the nonnegative rank of $\bm{\Phi}$ may be larger than the rank of $\bm{\Phi}$. However, if $\bm{\Phi}$ has rank 2, then the nonnegative rank is also 2 \citep{Thomas1974rank, de1991reduced, gillis2020nonnegative}. 

The identifiability issue for NMF is that for some $K \times K$ transformation matrix $\bm{S}$,
\begin{equation}\label{Eq:illustrationofuniquenessofnmf}
\bm{\Phi} = \bm{M}\bm{H} =  \left(\bm{M}\bm{S^{-1}}\right)\left(\bm{S}\bm{H}\right) = \tilde{\bm{M}}\tilde{\bm{H}}.
\end{equation}
The choice of $\bm{S}$ is restricted by the requirement that $\tilde{\bm{M}}$ and $\tilde{\bm{H}}$ should be nonnegative as well, just like $\bm{M}$ and $\bm{H}$. The identifiability issue will be discussed in more detail in Section~\ref{S: Identifiability issue}.

\subsection{LBA, EMA and asymmetric PLSA}\label{sub: LBA, EMA and PLSA}

LBA, EMA, and asymmetric PLSA have the same representation \citep{de1990latent, weltje1997end, hofmann2001unsupervised}. In LBA, EMA, and asymmetric PLSA, the observed matrix $\bm{P} = \bm{D}_{\bm{X}}^{-1}\bm{X}$ with row-sum-to-1 constraints is approximated by a lower rank matrix $\bm{\Pi}$, which is the product of two nonnegative matrices $\bm{W}$ and $\bm{G}$ with row-sum-to-1 constraints. Namely,
\begin{equation}\label{E: lbamod}
\begin{split}
    & \bm{\Pi} = \bm{W}\bm{G}\\
    \text{subject to} ~~~ &\bm{\Pi}\bm{1} = \bm{1}, \bm{W}\bm{1} = \bm{1}, \bm{G}\bm{1} = \bm{1},
\end{split}    
\end{equation}
\noindent where $\bm{W}$ and $\bm{G}$ have sizes $I\times K$ and $K\times J$ respectively, both have rank $K$, and $\bm{1}$ is the vector of all ones of appropriate dimension. The model decomposes the $I$ compositions in $\bm{\Pi}$ into $K$ latent compositions in $\bm{G}$, where $\bm{W}$ provides the mixing parameters that yield the $I$  compositions from the $K$ latent compositions. For convenience, we call $\bm{M}$ and $\bm{W}$ coefficient matrices and $\bm{H}$ and $\bm{G}$ basis matrices.

The identifiability issue for LBA, EMA and asymmetric PLSA is that for some $K \times K$ transformation matrix $\bm{T}$,
\begin{equation}\label{Eq: inllustration of uniqueness of lbaemaplsa}
\bm{\Pi} = \bm{W}\bm{G} =  \left(\bm{W}\bm{T}^{-1}\right)\left(\bm{T}\bm{G}\right) = \tilde{\bm{W}}\tilde{\bm{G}}.
\end{equation}
The choice of $\bm{T}$ is restricted by the requirement that $\tilde{\bm{W}}$ and $\tilde{\bm{G}}$ should be nonnegative and $\tilde{\bm{W}}\bm{1} = \bm{1},  \tilde{\bm{G}}\bm{1} = \bm{1}$, just like $\bm{W}$ and $\bm{G}$. The sum-to-1 constraints $\tilde{\bm{W}}\bm{1} = \bm{1},  \tilde{\bm{G}}\bm{1} = \bm{1}$ are implemented by imposing the constraint $\bm{T1} = \bm{1}$ on the elements of $\bm{T}$ \citep{de1990latent}. Thus weighted averages of the $K$ compositions in the $K \times J$ matrix $\bm{G}$, where the weights are in $\bm{T}$, lead to the new latent compositions in the
matrix $\tilde{\bm{G}}$, corresponding to the new weights for $\bm{\Pi}$ being in $\tilde{\bm{W}}$. The identifiability issue will be discussed in more details in Section~\ref{S: Identifiability issue}.

\subsection{LCA of a two-way table and symmetric PLSA}

LCA of a two-way table and symmetric PLSA are equivalent tools that exist under different names, where LCA is more often used in statistics and symmetric PLSA in computer sciences \citep{goodman1987new, Fienberg_Hersh_Rinaldo_Zhou_2009, hofmann1999probabilistic, hofmann2001unsupervised}. Let $\bm{Y}$ be an observed matrix with elements $x_{ij} / \sum_{ij}x_{ij}$ such that the sum of all elements in $\bm{Y}$ is 1. LCA of a two-way table and symmetric PLSA approximate $\bm{Y}$ by a lower rank matrix $\bm{\Psi}$ which is the product of three nonnegative matrices with sum-to-one constraints. The elements of $\bm{\Psi}$ can be interpreted as probabilities. The model is  
\begin{equation}\label{E: lcamod}
\begin{split}
    &  \bm{\Psi} = \bm{A}\Theta\bm{B}\\
    \text{subject to} ~~~ 
    &\bm{1}^T\bm{\Psi}\bm{1} = 1, \bm{A}^T\bm{1} = \bm{1}, \bm{1}^T\Theta\bm{1} = 1, \bm{B}\bm{1} = \bm{1}.
\end{split}    
\end{equation}
\noindent where $\Theta$ is a diagonal matrix, and $\bm{A}$, $\Theta$, and $\bm{B}$ are nonnegative matrices of sizes $I\times K$, $K\times K$, and $K\times J$, respectively. Due to the sum-to-1 constraints the elements of these matrices can be interpreted as (conditional) probabilities.

\subsection{Relations}

We first discuss the equivalence of LBA, EMA and asymmetric PLSA with LCA of a two-way table and symmetric PLSA \citep{van1989latent, heijden1994end, hofmann1999probabilistic}.
\begin{theorem}\label{theorem: equivalence}
LCA of a two-way table and symmetric PLSA are equivalent to LBA, EMA, and asymmetric PLSA: LCA of a two-way table and symmetric PLSA can be reparametrized into LBA, EMA, and asymmetric PLSA, and vice versa.
\end{theorem}
\begin{proof}
From (\ref{E: lbamod}) we have  $\bm{\Pi} = \bm{W}\bm{G}$, with $\bm{\Pi}\bm{1} = \bm{1}, \bm{W}\bm{1} = \bm{1}, \bm{G}\bm{1} = \bm{1}$. From (\ref{E: lcamod}) we have  $\bm{\Psi} = \bm{A}\Theta\bm{B}$, with $\bm{1}^T\bm{\Psi}\bm{1} = 1, \bm{A}^T\bm{1} = \bm{1}, \bm{1}^T\Theta\bm{1} = 1, \bm{B}\bm{1} = \bm{1}$.

First, suppose that we have LCA of a two-way table, symmetric PLSA of $\Psi$: $\bm{\Psi} = \bm{A}\Theta\bm{B}$, with $\bm{1}^T\bm{\Psi}\bm{1} = 1, \bm{A}^T\bm{1} = \bm{1}, \bm{1}^T\Theta\bm{1} = 1, \bm{B}\bm{1} = \bm{1}$. We have $\bm{D}_{\bm{\Psi}}^{-1}\bm{\Psi} = \bm{\Pi}$ as $\bm{D}_{\bm{\Psi}}^{-1}\bm{\Psi}\bm{1} = \bm{1}$. Then $\bm{\Pi} = \bm{D}_{\bm{\Psi}}^{-1}\bm{A}\Theta\bm{B}$.
Let $\bm{D}_{\bm{\Psi}}^{-1}\bm{A}\Theta = \bm{W}$ and $\bm{B} = \bm{G}$. We obtain LBA, EMA, asymmetric PLSA of $\bm{\Pi}$ as $\bm{D}_{\bm{\Psi}}^{-1}\bm{A}\Theta\bm{1} = \bm{D}_{\bm{\Psi}}^{-1}\bm{A}\Theta\bm{B}\bm{1}
= \bm{D}_{\bm{\Psi}}^{-1}\bm{\Psi}\bm{1} = \bm{1}$ and $\bm{B1} = \bm{1}$.

Second, suppose that we have LBA, EMA, asymmetric PLSA of $\bm{\Pi}$: $\bm{\Pi} = \bm{W}\bm{G}$. We have $\bm{D}_{\bm{\Psi}}\bm{\Pi} = \bm{\Psi}$ as $\bm{1}^T\bm{D}_{\bm{\Psi}}\bm{\Pi}\bm{1} = 1$. Then $\bm{\Psi} = \bm{D}_{\bm{\Psi}}\bm{W}\bm{G}$.
Let $\bm{D}_{\bm{\Psi}}\bm{W}(\text{diag}(\bm{1}^T\bm{D}_{\bm{\Psi}}\bm{W}))^{-1} = \bm{A}$, $\text{diag}(\bm{1}^T\bm{D}_{\bm{\Psi}}\bm{W}) = \Theta$, and $\bm{G} = \bm{B}$. We obtain LCA of a two-way table and symmetric PLSA of $\bm{\Psi}$ as $(\bm{D}_{\bm{\Psi}}\bm{W}(\text{diag}(\bm{1}^T\bm{D}_{\bm{\Psi}}\bm{W}))^{-1})^T\bm{1} = \bm{1}$, $\bm{1}^T\text{diag}(\bm{1}^T\bm{D}_{\bm{\Psi}}\bm{W})\bm{1} = 1$, and $\bm{G1} = \bm{1}$.
\end{proof}

In the following we will compare NMF only with LBA, EMA, and asymmetric PLSA because the results will also hold for LCA of a two-way table and symmetric PLSA. We have

\begin{theorem}\label{theorem: lbaplsanmfspecialcase}
LBA, EMA, and asymmetric PLSA are special cases of NMF, as $\bm{\Phi}$, $\bm{M}$, and $\bm{H}$ in NMF are nonnegative and further unrestricted, whereas $\bm{\Pi}$, $\bm{W}$, and $\bm{G}$ in LBA, EMA and asymmetric PLSA are nonnegative and have additional constraints $\bm{\Pi}\bm{1} = \bm{1}$, $\bm{W}\bm{1} = \bm{1}$, and $\bm{G}\bm{1} = \bm{1}$.
\end{theorem}
\begin{proof}
From (\ref{E: nmfmod}) we have NMF: $\bm{\Phi} = \bm{M}\bm{H}$. From (\ref{E: lbamod}) we have LBA, EMA, asymmetric PLSA: $\bm{\Pi} = \bm{W}\bm{G}$, with $\bm{\Pi}\bm{1} = \bm{1}, \bm{W}\bm{1} = \bm{1}, \bm{G}\bm{1} = \bm{1}$. 
\end{proof}

Disregarding the objective function and algorithm, a decomposition by NMF and a decomposition by LBA, EMA and asymmetric PLSA are related as follows  \citep{de1991reduced}: 
\begin{theorem}\label{theorem: lbaplsanmfsolution}
NMF of $\bm{\Phi}$ can be transformed into LBA, EMA, and asymmetric PLSA of $\bm{\Pi}$; the other way around, using $\bm{D}_{\bm{\Phi}}$, LBA, EMA, and asymmetric PLSA of $\bm{\Pi}$ can be transformed into NMF of $\bm{\Phi}$.
\end{theorem}
\begin{proof}
Suppose that ($\bm{M}$, $\bm{H}$) is NMF of $\bm{\Phi}$. We have $\bm{D}_{\bm{\Phi}}^{-1}\bm{\Phi} = \bm{\Pi}$ as $\bm{D}_{\bm{\Phi}}^{-1}\bm{\Phi}\bm{1} = \bm{1}$. Then $\bm{\Pi} = \bm{D}_{\bm{\Phi}}^{-1}\bm{M}\bm{H}$ where $\bm{D}_{\bm{H}}^{-1}\bm{H} = \bm{G}$ and $\bm{D}_{\bm{\Phi}}^{-1}\bm{M}\bm{D}_{\bm{H}} = \bm{W}$ as $\bm{D}_{\bm{H}}^{-1}\bm{H}{1} = \bm{1}$ and $\bm{D}_{\bm{\Phi}}^{-1}\bm{M}\bm{D}_{\bm{H}}\bm{1} = \bm{D}_{\bm{\Phi}}^{-1}\bm{M}\bm{D}_{\bm{H}}(\bm{D}_{\bm{H}}^{-1}\bm{H})\bm{1}
= \bm{D}_{\bm{\Phi}}^{-1}\bm{M}\bm{H}\bm{1} = \bm{1}$. 

Suppose we have LBA, EMA, asymmetric PLSA of $\bm{\Pi}$: $\bm{\Pi} = \bm{W}\bm{G}$, and $\bm{D}_{\bm{\Phi}}$ is known. We have $\bm{D}_{\bm{\Phi}}\bm{\Pi} = \bm{\Phi}$. Then $\bm{\Phi} = \bm{D}_{\bm{\Phi}}\bm{W}\bm{G}$ where $\bm{D}_{\bm{\Phi}}\bm{W} = \bm{M}$ and $\bm{G} = \bm{H}$. We obtain NMF of $\bm{\Phi}$ as $\bm{D}_{\bm{\Phi}}\bm{W}$ and $\bm{G}$ are nonnegative.
\end{proof}

To the best of our knowledge, our work appears to be the first to link LBA from social science and EMA from geology to PLSA and NMF from computing science.

\section{Identifiability issue}\label{S: Identifiability issue}

\noindent In this section, first we give an introduction to the identifiability issue in Section~\ref{Sub: introduction}. Then we give a review of the sufficient conditions for the solutions of LBA, EMA, PLSA to be unique in Section~\ref{Sub: identlba} and for the solution of NMF to be unique in Section~\ref{Sub: sufnmf}. In Section~\ref{Sub: ifandonlyif} we provide a theorem showing that uniqueness of NMF  holds if and only if uniqueness of LBA, EMA, and asymmetric PLSA hold. Making use of this theorem, the identifiability theorem in NMF is also suitable for LBA, PLSA, and EMA, and vice versa. Finally, we provide a conclusion in Section~\ref{Sub: condis}.

\subsection{Introduction}\label{Sub: introduction}

In the following, we will use results on convex cones and convex hulls \citep{sandgren1954convex, avis1995good, gillis2020nonnegative}. Given a matrix $\bm{U}$ of size $I\times J$, the convex cone of $\bm{U}$, denoted by $\text{cone}(\bm{U})$, is the convex cone generated by the columns of $\bm{U}$: $\bm{U}(:, j)$ ($j = 1, \cdots, J$). That is, for any nonnegative weights $r_j \geq 0$ ($j = 1, \cdots, J$), the weighted sum $\sum_jr_j\bm{U}(:, j)\in \text{cone}(\bm{U})$. It can be expressed as 
\begin{equation*}
\text{cone}(\bm{U}) = \{\bm{u} | \bm{u} = \sum_{j = 1}^{J}r_j\bm{U}(:, j), r_j \geq 0, j = 1, 2, \cdots, J\}.
\end{equation*}
The convex hull of $\bm{U}$, denoted by $\text{conv}(\bm{U})$, is the convex hull generated by the columns of $\bm{U}$. It can be expressed as follows:
\begin{equation*}
    \text{conv}(\bm{U}) = \{\bm{u} | \bm{u} = \sum_{j = 1}^{J}r_j\bm{U}(:, j), \sum_{j = 1}^{J} r_j = 1, r_j \geq 0, j = 1, 2, \cdots, J\}.
\end{equation*}
The difference between $\text{conv}(\bm{U})$ and $\text{cone}(\bm{U})$ is that $\text{conv}(\bm{U})$ adds the sum-to-one constraint on weights $r_j$ ($j = 1, \cdots, J$): $\sum_{j = 1}^Jr_j = 1$. If $\bm{U}(:, 1), \bm{U}(:, 2), ..., \bm{U}(:, J)$ are affine independent, $\text{conv}(\bm{U})$ is the so-called $(J-1)$-dimensional simplex and $\bm{U}(:, 1), \bm{U}(:, 2), ..., \bm{U}(:, J)$ are so-called vertices of the simplex.

Before we give a definition of identifiability, two toy examples are provided to illustrate geometrically that LBA, EMA, asymmetric PLSA, and NMF are  non-unique. Table~\ref{T: healthgenderoriginal} is a contingency table of size $5\times 2$, where rows represent self-assessed health and columns represent two genders \citep{greenacre2017correspondence}; Table~\ref{T: educationreadshiporiginal} is a contingency table of size $5\times 3$ where rows represent education groups and columns represent readship class \citep{greenacre2017correspondence}. Table~\ref{T: healthgendersumto1} and Table~\ref{T: educationreadshipsumto1} are normalized versions of Table~\ref{T: healthgenderoriginal} and Table~\ref{T: educationreadshiporiginal} respectively, where elements in each row sum to 1.

\begin{table}[h]
\centering  
\caption{(a) Contingency table of self-assessed health with gender \citep{greenacre2017correspondence}; (b) Contingency table of education group by readship class, where E1, E2, E3, E4, and E5 denote Some primary, Primary completed, Some secondary, Secondary completed, and Some tertiary, respectively, and C1, C2, and C3 denote Glance, Fairly thorough, and Very thorough, respectively \citep{greenacre2017correspondence}.} 
\label{T: healthgender}
  \begin{subtable}[t]{0.30\textwidth}
  \caption{Data source: Exhibit 16.1 \citep{greenacre2017correspondence}} 
  \label{T: healthgenderoriginal}
    \centering
\begin{tabular}{lrrllll}
\hline
 & Male & Female \\ 
  \hline
Very good & 448 & 369 \\ 
  Good & 1789 & 1753 \\ 
  Regular & 636 & 859 \\ 
  Bad & 177 & 237 \\ 
  Very bad & 39 & 64 \\ 
   \hline
\end{tabular} 
 \end{subtable}
 \hspace{2cm}
   \begin{subtable}[t]{0.30\textwidth}
  \caption{ Data source: Exhibit 3.1 \citep{greenacre2017correspondence}} 
  \label{T: educationreadshiporiginal}
    \centering
\begin{tabular}{lrrrlll}
\hline
 & C1 & C2 & C3 \\ 
  \hline
  E1 & 5 & 7 & 2 \\ 
  E2 & 18 & 46 & 20 \\ 
  E3 & 19 & 29 & 39 \\ 
  E4 & 12 & 40 & 49 \\ 
  E5 & 3 & 7 & 16 \\ 
   \hline
\end{tabular} 
 \end{subtable}
\end{table}

\begin{table}[h]
\centering  
\caption{(a) Normalized Table~\ref{T: healthgenderoriginal} such that each row is sum-to-1; (b) Normalized Table~\ref{T: educationreadshiporiginal} such that each row is sum-to-1} 
\label{T: educationreadship}
 \begin{subtable}[t]{0.30\textwidth}
\caption{Normalized Table~\ref{T: healthgenderoriginal}} 
\label{T: healthgendersumto1}
\centering  
\begin{tabular}{lrrllll}
    \hline
 & Male & Female \\ 
  \hline
very good & 0.548 & 0.452 \\ 
  good & 0.505 & 0.495 \\ 
  regular & 0.425 & 0.575 \\ 
  bad & 0.428 & 0.572 \\ 
  very bad & 0.379 & 0.621 \\ 
   \hline
\end{tabular}
\end{subtable}
 \hspace{2cm}
 \begin{subtable}[t]{0.30\textwidth}
\caption{Normalized Table~\ref{T: educationreadshiporiginal}} 
\label{T: educationreadshipsumto1}
\centering  
\begin{tabular}{lrrrlll}
   \hline
 & C1 & C2 & C3 \\ 
  \hline
E1 & 0.357 & 0.500 & 0.143 \\ 
  E2 & 0.214 & 0.548 & 0.238 \\ 
  E3 & 0.218 & 0.333 & 0.448 \\ 
  E4 & 0.119 & 0.396 & 0.485 \\ 
  E5 & 0.115 & 0.269 & 0.615 \\ 
   \hline
\end{tabular}
\end{subtable}
\end{table}

\begin{figure}[h]
\centering
  \begin{subfigure}[b]{0.3\linewidth}
 \includegraphics[width=1\textwidth]{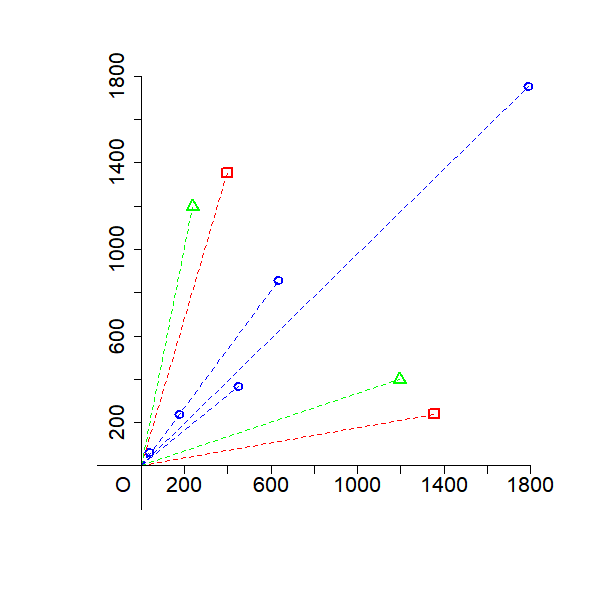}
 \caption{Table~\ref{T: healthgenderoriginal}}\label{F: greenacrep122dim2}
 \end{subfigure}
   \begin{subfigure}[b]{0.3\linewidth}
 \includegraphics[width=1\textwidth]{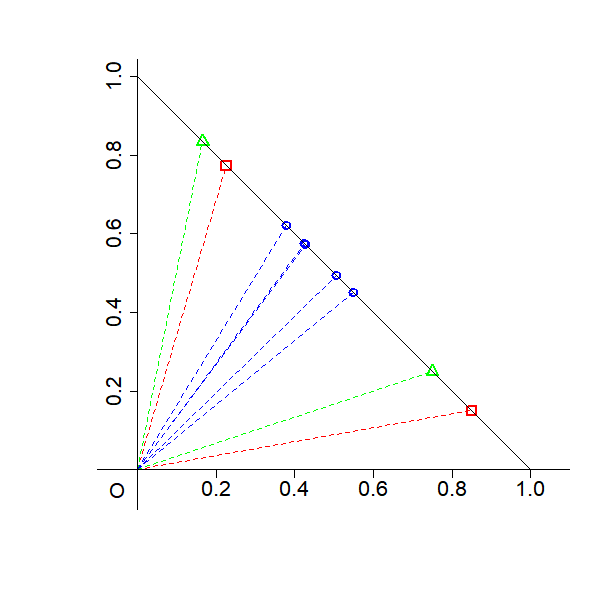}
    \caption{Table~\ref{T: healthgendersumto1}}
    \label{F: greenacrep122dim2l1}
 \end{subfigure}
    \begin{subfigure}[b]{0.3\linewidth}
 \includegraphics[width=1\textwidth]{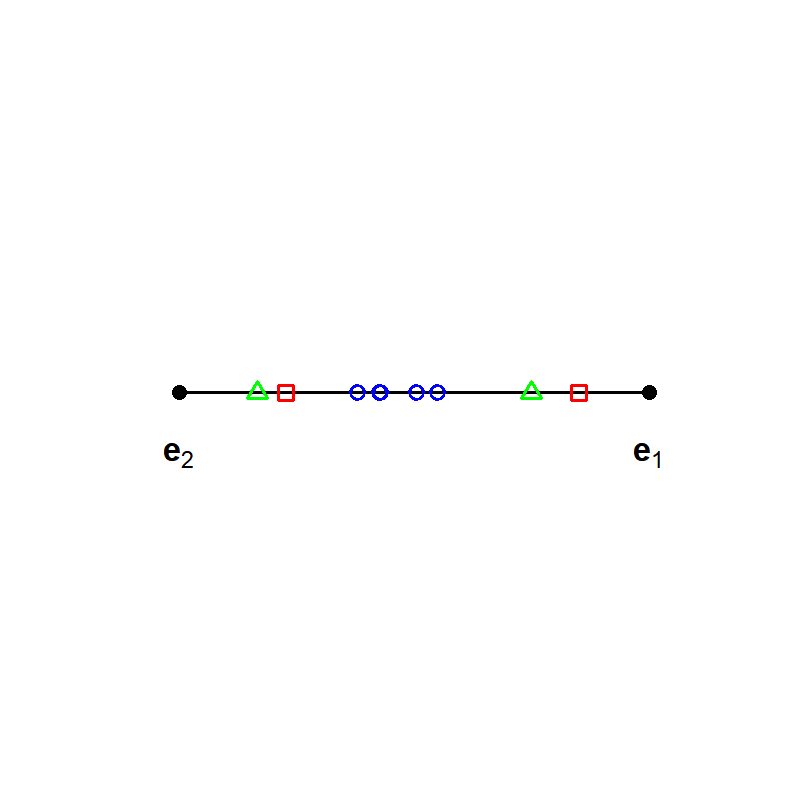}
    \caption{Taken out from Figure~\ref{F: greenacrep122dim2l1}}
    \label{F: greenacrep122dim2simplex}
 \end{subfigure}
 \caption{Geometric illustration that the solutions of NMF, LBA, EMA, and asymmetric PLSA are not unique using Table~\ref{T: healthgenderoriginal} and Table~\ref{T: healthgendersumto1} where rows in Table~\ref{T: healthgenderoriginal} and Table~\ref{T: healthgendersumto1} are in blue and two possible basis matrices are in green and in red.}\label{F: greenacrep122dim2originaldata}
    \end{figure}

\begin{figure}
\centering
  \begin{subfigure}[b]{0.3\linewidth} \includegraphics[width=1\textwidth]{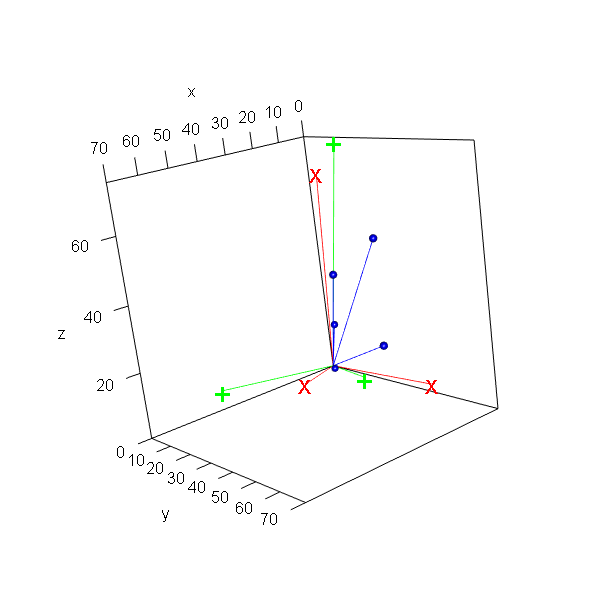}
 \caption{Table~\ref{T: educationreadshiporiginal}}\label{F: greenacrep18dim3}
 \end{subfigure}
   \begin{subfigure}[b]{0.3\linewidth}
 \includegraphics[width=1\textwidth]{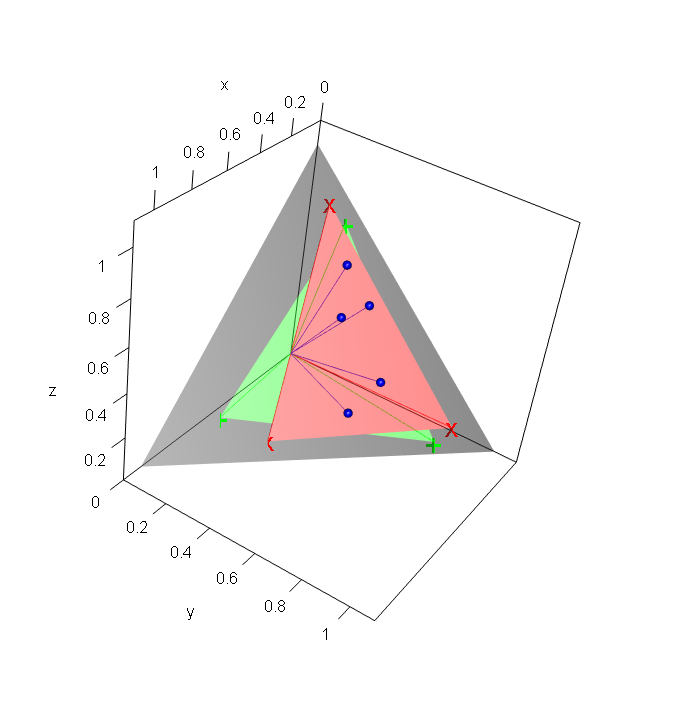}
    \caption{Table~\ref{T: educationreadshipsumto1}}
    \label{F: greenacrep18dim3l1}
 \end{subfigure}
     \begin{subfigure}[b]{0.3\linewidth}
 \includegraphics[width=1\textwidth]{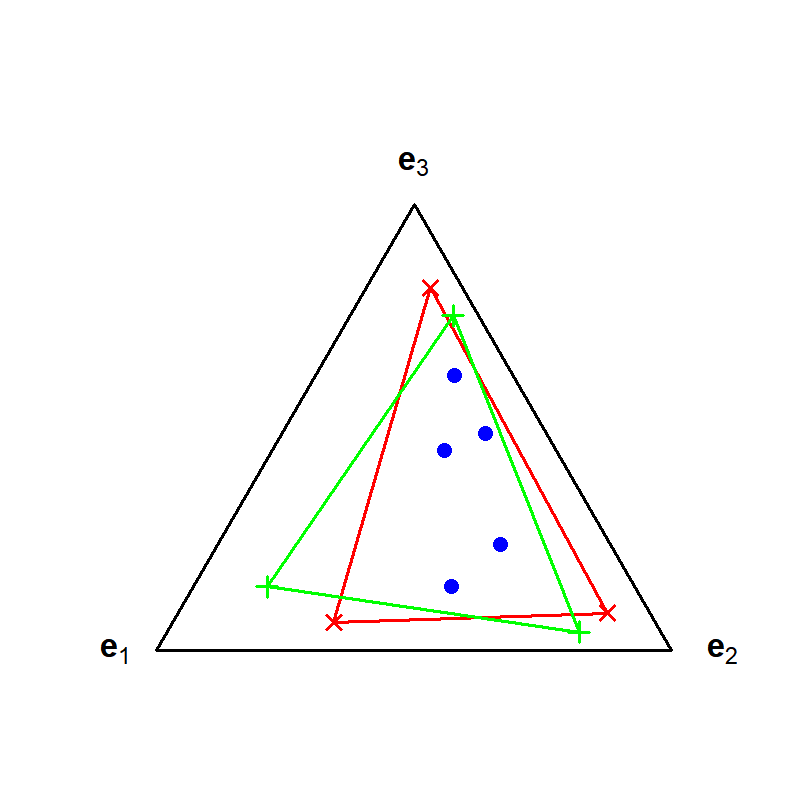}
    \caption{Taken out from Figure~\ref{F: greenacrep18dim3l1}}
    \label{F: greenacrep18dim3simplex}
 \end{subfigure}
 \caption{Geometric illustration that the solutions of NMF, LBA, EMA, and asymmetric PLSA are not unique using Table~\ref{T: educationreadshiporiginal} and Table~\ref{T: educationreadshipsumto1} where rows in Table~\ref{T: educationreadshiporiginal} and Table~\ref{T: educationreadshipsumto1} are in blue and two possible basis matrices are in green and in red.}\label{F: greenacrep18dim3originaldata}
    \end{figure}

The NMF decomposition $\bm{\Phi} = \bm{MH}$ means that each row of $\bm{\Phi}$ is a linear combination of the rows of $\bm{H}$ weighted by the components of the corresponding row of $\bm{M}$ \citep{berman1994nonnegative, gillis2020nonnegative}. This provides a geometric interpretation to NMF: the rows of $\bm{\Phi}$ are inside the convex cone generated by the rows of $\bm{H}$. Rows of Table~\ref{T: healthgenderoriginal} and Table~\ref{T: educationreadshiporiginal} are respectively presented in Figure~\ref{F: greenacrep122dim2} and Figure~\ref{F: greenacrep18dim3} as blue points. Both points in red and points in green could serve as two different basis matrices $\bm{H}$. Thus the NMF solution is not unique.

Likewise, in LBA, EMA, and asymmetric PLSA, $\bm{\Pi} = \bm{WG}$ means that geometrically the rows of $\bm{\Pi}$ are contained in the convex hull generated by the rows of $\bm{G}$. See Figure~\ref{F: greenacrep122dim2l1}
and Figure~\ref{F: greenacrep18dim3l1} for Table~\ref{T: healthgendersumto1} and Table~\ref{T: educationreadshipsumto1}, respectively.
Figure~\ref{F: greenacrep122dim2simplex} and Figure~\ref{F: greenacrep18dim3simplex} are the simplexes  taken out from Figure~\ref{F: greenacrep122dim2l1}
and Figure~\ref{F: greenacrep18dim3l1} respectively, which shows that the sum-to-one restriction for each row reduces the dimensionality with one. Both red points and green points are possible basis matrices $\bm{G}$, demonstrating that the LBA, EMA, and asymmetric PLSA solution is not unique.

The identifiability issue of NMF, LBA, EMA, and asymmetric PLSA is also known as the nonuniqueness issue \citep{gillis2020nonnegative}. The uniqueness of a NMF decomposition is formally defined as follows  \citep{huang2014non, gillis2020nonnegative}.
\begin{definition}\label{definition: theorem4.5gillis}
    The NMF solution ($\bm{M}$, $\bm{H}$) of $\bm{\Phi}$ is said to be unique if and only if any other NMF solution ($\tilde{\bm{M}}$, $\tilde{\bm{H}}$) has the form
    \begin{equation*}
        \tilde{\bm{M}} = \bm{M}(\bm{\Gamma}\bm{\Sigma})^{-1} \text{ and } \tilde{\bm{H}} = (\bm{\Gamma}\bm{\Sigma})\bm{M}
    \end{equation*}
    where $\bm{\Gamma}$ is a permutation matrix and $\bm{\Sigma}$ is a diagonal matrix whose diagonal entries are positive.
\end{definition}
\noindent The same definition is given for the solution ($\bm{W}$, $\bm{G}$) of LBA, EMA, and asymmetric PLSA to be unique, except that the diagonal matrix $\bm{\Sigma}$ can be omitted. As mentioned in Section~\ref{sub: LBA, EMA and PLSA}, this is because each row of the transformation matrix $\bm{T}$ is constrained to sum to one \citep{de1990latent}. 
\begin{definition}\label{definition: de1990latent}
    The LBA, EMA, and asymmetric PLSA solution ($\bm{W}$, $\bm{G}$) of $\bm{\Pi}$ is said to be unique if and only if any other LBA, EMA, and asymmetric PLSA solution ($\tilde{\bm{W}}$, $\tilde{\bm{G}}$) has the form
    \begin{equation*}
        \tilde{\bm{W}} = \bm{W}\bm{\Gamma}^{-1} \text{ and } \tilde{\bm{G}} = \bm{\Gamma}\bm{G}
    \end{equation*}
    where $\bm{\Gamma}$ is a permutation matrix.
\end{definition}

When $K = 1$, NMF, LBA, EMA, and asymmetric PLSA are unique because the transformation matrix $\bm{S}$ in NMF is a scalar (see Eq. (\ref{Eq:illustrationofuniquenessofnmf})) and $\bm{T}$ in LBA, EMA, asymmetric PLSA is 1 (see Eq. (\ref{Eq: inllustration of uniqueness of lbaemaplsa})).

\subsection{Proposals for identifiability of LBA, EMA, PLSA}\label{Sub: identlba}

We have illustrated that, in general, the solution of asymmetric PLSA is not unique. However, this is not always
acknowledged in the PLSA literature, see, for instance, \citet{gaussier2005relation}.

In contrast, the identifiability issue in LBA is well recognized \citep{de1990latent, van1992constrained, van1999identifiability}.
A classic proposal aiming at addressing the identifiability issue involves the, what they call, inner extreme solution and outer extreme solution, which are determined by the transformation matrix $\bm{T}$: $(\bm{W}\bm{T}^{-1}, \bm{T}\bm{G})$. The inner extreme solution selects $\bm{T}$ such that the basis vectors in the rows of $\bm{T}\bm{G}$ are as similar as possible, while outer extreme solution selects $\bm{T}$ such that their difference is maximized. 

For $K = 2$, the transformation matrix $\bm{T}$ is characterized by two parameters $x$, $y$: $\bm{T} = \begin{bmatrix}x & 1-x \\y & 1-y\end{bmatrix}$ with feasible ranges for $(x, y)$ governed by nonnegative constraints in $\bm{W}$ and $\bm{G}$ \citep{de1990latent, moussaoui2005non}. The following theorem is given:
\begin{theorem}\label{theorem: lbak2}[\citep{de1990latent, van1999identifiability}]
   When $K = 2$, both inner and outer extreme solutions are unique. The solution is inner extreme if and only if  
   $\bm{W}$ contains a permuted $2\times 2$ identity matrix, while the
  solution is outer extreme if and only if $\bm{G}$ contains a permuted $2\times 2$ diagonal matrix.
\end{theorem}
\noindent This means that, for the inner extreme solution, two rows of $\bm{G}$ are picked from two rows of $\bm{\Pi}$ and thus the coefficient matrix $\bm{W}$ includes a permuted identity matrix. As an example, in Figure~\ref{F: greenacrep122dim2simplex} basis vectors of the inner extreme solution are the most extreme blue data point on the left and the most extreme blue data point on the right. In contrast, for the outer extreme solution, two rows of $\bm{G}$ are $\bm{e}_1 = (1, 0)$ and $\bm{e}_2 = (0, 1)$, i.e. the two black end points in Figure~\ref{F: greenacrep122dim2simplex}.

For $K > 2$, the terms “as similar as possible” and “as different as possible” have to be explicitly defined. To quantify similarity and dissimilarity between basis vectors from the rows of $\bm{G}$, three criteria were proposed in \citet{van1999identifiability}: (1) the sum of distances between the basis vectors, (2) the volume spanned by the basis vectors, (3) statistical dependence of basis vectors, where the first criterion was elaborated. They propose a Monte Carlo method to find a transformation matrix 
$\bm{T}$ such that the sum of chi-square distances between basis vectors 
\begin{equation}\label{E: lbamodchisquareinnerouter}
    \sum_{k = 1}^K\sum_{k' = k+1}^{K} \sqrt{\sum_{j = 1}^{J}\frac{(g_{kj}-g_{k'j})^2}{\sum_i\phi_{ij}}}
\end{equation}
is either minimized ("inner extreme solution") or maximized ("outer extreme solution"), where $g_{kj}$ denotes element ($k,j$) of $\bm{G}$ and $\phi_{ij}$ denotes element ($i,j$) of $\bm{\Phi}$. The chi-square distance, playing a central role in methods like correspondence analysis \citep{Qi_Hessen_Deoskar_vanderHeijden_2024}, weights the squared difference between basis vectors elements $(g_{kj}-g_{k'j})^2$ by the marginal proportion 
$\sum_i\phi_{ij}$. This adjustment corrects the difference $(g_{kj}-g_{k'j})^2$ for the size of column $j$ \citep{van1999identifiability}. 

For $K > 2$, inner or outer extreme solutions may be not unique. For example, for $K = 3$, multiple triangles formed by the rows of different basis matrices $\bm{G}$ could achieve the same minimum or maximum total chi-square distance. Figure~\ref{F: greenacrep18dim3simplex} illustrates this situation, where the basis matrix $\bm{G}$ corresponding to the red points is assumed to have the same total chi-square distance as the basis matrix $\bm{G}$ corresponding to the green points.

For EMA, minimum end members \citep{weltje1997end, weltje2007genetically, seidel2015r, renner1993resolution, renner1995construction} and outermost end members \citep{ZHANG2020106656} have been proposed as approaches aimed at achieving uniqueness, analogous to inner and outer extreme solutions in LBA, respectively. A sufficient condition for EMA with outermost end members to be unique was provided in \citet{qi2026unmixing}, drawing on results from minimum-volume NMF, as described in the next section. Note that, in EMA, basis vectors and coefficient vectors are usually called end members and abundances, respectively. 

In addition to the inner extreme solution and the outer extreme solution in LBA, imposing constraints on $\bm{W}$ and $\bm{G}$ was also discussed in \citet{van1992constrained} to obtain uniqueness. Three types of constraints for LBA were suggested: fixed value, equality, and multinomial logit constraints. However, no sufficient conditions are provided to guarantee a unique solution. In this paper we ignore these constraints, but note that for $K = 2$ the inner extreme solution and outer extreme solution can be obtained by imposing zero-value constraints on $\bm{W}$ and $\bm{G}$, respectively. 

\subsection{Proposals for identifiability of NMF}\label{Sub: sufnmf}

The sufficient conditions for NMF to have a unique solution are well studied, where separability
and minimum volume
are two main assumptions \citep{gillis2020nonnegative}.  Note that many NMF work takes $\bm{M}$ as basis matrix and $\bm{H}$ as coefficient matrix. Here, to keep consistent with LBA, EMA, and asymmetric PLSA, we take $\bm{M}$ as coefficient matrix and $\bm{H}$ as basis matrix. This can be achieved by $\bm{\Phi}^T = \bm{H}^T\bm{M}^T$.

Separability is first introduced by \citet{NIPS2003_1843e35d}. A matrix $\bm{M}$ of size $I\times K$ is said to be separable or have a separability property if it contains a $K \times K$ submatrix that is a permutation of a diagonal matrix with positive
diagonal entries \citep{gillis2020nonnegative}. The uniqueness theorem of NMF under the separability assumption can be stated as follows:
\begin{theorem}\label{theorem: separability}
\citep{Chan2009Convex, Zhou6003792, gillis2020nonnegative} Suppose that $\text{rank}(\bm{\Phi}) = \text{rank}(\bm{M}) = \text{rank}(\bm{H})$ and the coefficient matrix $\bm{M}$ is separable. Then the NMF solution ($\bm{M}$, $\bm{H}$) of $\bm{\Phi}$ is unique under these assumptions.
\end{theorem}
\noindent Separability on $\bm{M}$ 
means that basis vectors (from the rows of $\bm{H}$) are themselves data points (from the rows of $\bm{\Phi}$) up to scaling. 

When $K = 2$, Theorem~\ref{theorem: lbak2} for the inner extreme solution of LBA coincides with Theorem~\ref{theorem: separability}, but Theorem~\ref{theorem: separability} considers uniqueness for any dimensionality $K$. Note that in LBA, EMA, and asymmetric PLSA, due to $\bm{W1} = \bm{1}$, separability on $\bm{W}$ corresponds to containing an identity matrix of size $K \times K$ up to permutation.

\begin{figure}
\centering
  \begin{subfigure}[b]{0.45\linewidth}
\includegraphics[width=1\textwidth]{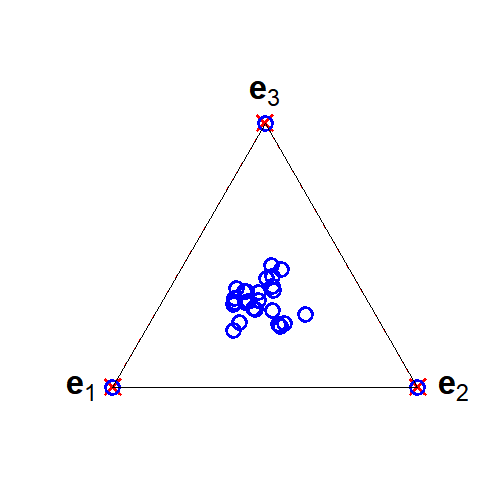}
 \caption{$\bm{M}$ satisfies separability}\label{F: separability}
 \end{subfigure}
   \begin{subfigure}[b]{0.45\linewidth}
\includegraphics[width=1\textwidth]{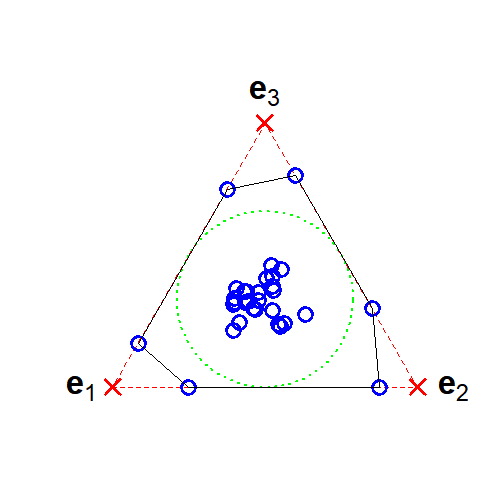}
    \caption{$\bm{M}^T$ satisfies SSC}
    \label{F: halfpurity}
 \end{subfigure}
 \caption{Geometric illustrations of (a) separability of a matrix $\bm{M}$ and (b) SSC of a matrix $\bm{M}^T$, obtained by assuming that the viewer stands in the nonnegative orthant, faces the origin, and looks at the two-dimensional plane $\bm{x1} = \bm{1}$ \citep{gillis2020nonnegative}. The blue dots "o" represent the rows of $\bm{M}$, and the red crosses "X" denote the standard basis vectors $\bm{e}_1, \bm{e}_2$, $\bm{e}_3$.}\label{F:      sepapurityssc}
    \end{figure}

See Figure~\ref{F: separability} for an illustration that a coefficient matrix $\bm{M}$ satisfies the separability assumption for $K = 3$ \citep{gillis2020nonnegative}. Here, $\bm{M}$ has size of $28\times 3$, meaning that it consists of 28 data points, each represented as a three-dimensional vector. Assuming that the viewer stands in the nonnegative orthant, faces the origin, and looks at the two-dimensional plane $\bm{x1} = \bm{1}$ \citep{gillis2020nonnegative}, we obtain Figure~\ref{F: separability}. In the figure, the blue dots "o" represent the rows of $\bm{M}$, and the red crosses "X" denote the standard basis vectors $\bm{e}_1, \bm{e}_2$, $\bm{e}_3$, where $\bm{e}_k$ is the three-dimensional standard basis vector whose $k$th element is 1 and all other elements are 0. The triangle corresponds to the nonnegative orthant cone $\text{cone}(\bm{e}_1, \bm{e}_2, \bm{e}_3)$, and the polygon formed by the dots is $\text{cone}(\bm{M}^T)$. From the figure, we can see that $\text{cone}(\bm{M}^T)$ coincides with $\text{cone}(\bm{e}_1, \bm{e}_2, \bm{e}_3)$. For every $k \in \{1, 2, 3\}$, there exists a row $i \in \{1, \cdots, I\}$, such that $\bm{m}_i = \alpha_k\bm{e}_k$ where $\bm{m}_i$ is $i$th row of $\bm{M}$ and $\alpha_k$ is a scalar. Theoretically appealing as it may be, this approach may fail to hold in applications like text mining, where basis vectors are not equal to existing data points. 

The minimum volume assumption remains robust when the separability assumption is violated. Under the minimum volume assumption, the determinant
\begin{equation}\label{Eq: minvolnmfmodonly}
\text{det}(\bm{H}\bm{H}^T)
\end{equation}
is minimal \citep{lin2015identifiability, fu2015}. The rationale for using $\text{det}(\bm{H}\bm{H}^T)$ as the objective function in Eq.~(\ref{Eq: minvolnmfmodonly}) is that $\sqrt{\text{det}(\bm{H}\bm{H}^T)}/K!$ corresponds to the volume of the simplex $\text{conv}(\bm{h}_1, \cdots, \bm{h}_K)$ and the origin, where $\bm{h}_k$ denotes the row $k$ from the basis matrix $\bm{H}$ \citep{leplat2020}.

The minimum volume assumption in NMF is analogous to the inner extreme solution in LBA \citep{lin2015identifiability, fu2015, van1999identifiability}: the former seeks basis vectors (i.e., rows of $\bm{H}$) whose convex hull encloses all data points (i.e., rows of $\bm{\Phi}$) with minimum volume, whereas the latter seeks basis vectors (i.e., rows of $\bm{G}$) whose convex hull encloses all data points (i.e., rows of $\bm{\Pi}$) while minimizing the total chi-square distance among them. 

Under the minimum volume assumption, several sufficient conditions have been proposed to ensure that $\bm{\Phi}$ has a unique NMF solution ($\bm{M}$, $\bm{H}$) \citep{lin2015identifiability, fu2015, fu2018identifiability, leplat2020}. The following theorem, based on the so-called sufficiently scattered condition (SSC), was provided in \citet{fu2015}.
\begin{theorem}\label{theorem: futheorem1}
[\citet{fu2015}] Suppose that $\text{rank}(\bm{\Phi}) = \text{rank}(\bm{M}) = \text{rank}(\bm{H})$ and the minimum volume assumption holds. If $\bm{M}\bm{1} = \bm{1}$ and $\bm{M}^T$ satisfies the so-called sufficiently scattered condition (SSC),
the NMF solution ($\bm{M}$, $\bm{H}$) of $\bm{\Phi}$ is unique under these assumptions.
\end{theorem}
\noindent A matrix $\bm{M}^T$ is sufficient scattered if (1) SSC1: $\mathbb{C} \subseteq \text{cone}(\bm{M}^T)$, where $\mathbb{C} = \{x \in \Re_+^K | \bm{1}^T\bm{x} \geq \sqrt{K - 1}||x||_2\}$ is second-order cone, and (2) SSC2: there does not exist any orthogonal matrix $\bm{Q} \in \Re^{K\times K}$ such that cone($\bm{M}^T$) $\subseteq$ cone($\bm{Q}$) except for permutation matrices. \citet{lin2015identifiability} proposed another sufficient condition for NMF to be unique under $\text{rank}(\bm{\Phi}) = \text{rank}(\bm{M}) = \text{rank}(\bm{H})$ and the minimum volume assumption. \citet{fu2016robust} later proved that this condition is equivalent to the one in Theorem~\ref{theorem: futheorem1}. 

Figure~\ref{F: halfpurity} serves as an illustration that a matrix $\bm{M}^T$ satisfies SSC for $K = 3$ \citep{gillis2020nonnegative}. Here, $\bm{M}$ has size of $31\times 3$.
The triangle corresponds to the nonnegative orthant cone $\text{cone}(\bm{e}_1, \bm{e}_2, \bm{e}_3)$, the polygon formed by the dots is $\text{cone}(\bm{M}^T)$, and the circle is the second-order cone $\mathbb{C}$. From the figure, we can see that the circle is contained in the polygon formed by the dots. Thus SSC1 holds. Any orthogonal matrix is a rotated version of the triangle in the figure. We can see that no rotated version of this triangle (except for itself) can contain the polygon formed by the dots. Thus SSC2 hold.

Theorem~\ref{theorem: futheorem1} has the condition that the sum of each row of coefficient matrix $\bm{M}$ is 1: $\mathbf{M}\mathbf{1} = \mathbf{1}$. Instead of $\mathbf{M}\mathbf{1} = \mathbf{1}$, \citet{fu2018identifiability} extended this result to the case where $\mathbf{M}^T\mathbf{1} = \mathbf{1}$; \citet{leplat2020} proved the case where $\mathbf{H}\mathbf{1} = \mathbf{1}$. In addition, \citet{qi2026identification} provided maximum-volume NMF, under SSC on basis matrix $\bm{H}$, which is similar with \citet{qi2026unmixing}.

Beyond separability, minimum volume, and maximum volume approaches, other frameworks have been proposed to ensure the uniqueness of NMF solutions such as sparsity, orthogonality, sub-separability, among others \citep{gillis2012sparse, ge2015intersecting, Lin2018maximum, gillis2020nonnegative, javadi2020nonnegative, abdolali2021simplex, Gillis2023, Abdolali2024Dual}. Necessary conditions have also been studied, including those in \citet{moussaoui2005non, huang2014non}.

\subsection{NMF uniqueness iff LBA, EMA, and asymmetric PLSA uniqueness}\label{Sub: ifandonlyif}

We now discuss how uniqueness of NMF is related to uniqueness of LBA, EMA, and asymmetric PLSA. We provide the following theorem:
\begin{theorem}\label{theorem: mainresultsnonunique}
The solution of NMF of $\bm{\Phi}$ is not unique if and only if the solution of LBA, EMA, and asymmetric PLSA of $\bm{\Pi}$ is not unique.
\end{theorem}
\begin{proof}
    Suppose that the solution of NMF of $\bm{\Phi}$ is not unique. Thus $\bm{S}$ is not simply a permutation matrix, see Definition~\ref{definition: theorem4.5gillis}. We have $\bm{\Phi} = \bm{M}\bm{H} = \left(\bm{M}\bm{S}^{-1}\right)\left(\bm{S}\bm{H}\right) = \tilde{\bm{M}}\tilde{\bm{H}}$, as in Eq.~(\ref{Eq:illustrationofuniquenessofnmf}). From Theorem~\ref{theorem: lbaplsanmfsolution}, both $(\bm{W}, \bm{G})$ and $(\tilde{\bm{W}}, \tilde{\bm{G}})$ are solutions of LBA, EMA, and asymmetric PLSA of $\bm{\Pi}$, where $\bm{W} = \bm{D}_{\bm{\Phi}}^{-1}\bm{M}\bm{D}_{\bm{H}}$, $\bm{G} = \bm{D}_{\bm{H}}^{-1}\bm{H}$, $\tilde{\bm{W}} = \bm{D}_{\bm{\Phi}}^{-1}(\bm{M}\bm{S}^{-1})\bm{D}_{(\bm{SH})}$, and $\tilde{\bm{G}} = \bm{D}_{(\bm{SH})}^{-1}(\bm{SH})$. Thus, we have $\tilde{\bm{W}} = \bm{W}(\bm{D}_{\bm{H}}^{-1}\bm{S}^{-1}\bm{D}_{(\bm{SH})})$ and $\tilde{\bm{G}} = (\bm{D}_{(\bm{SH})}^{-1}\bm{S}\bm{D}_{\bm{H}})\bm{G}$. Because $\bm{D}_{(\bm{SH})}^{-1}\bm{S}\bm{D}_{\bm{H}}$ is not a permutation matrix, the solution of LBA, EMA, and asymmetric PLSA of $\bm{\Pi}$ is not unique. 
    
    Suppose that the solution of LBA, EMA, and asymmetric PLSA of $\bm{\Pi}$ is not unique. Thus $\bm{T}$ is not simply a permutation matrix, see Definition~\ref{definition: de1990latent}. We have $\bm{\Pi} = \bm{W}\bm{G} = \left(\bm{W}\bm{T}^{-1}\right)\left(\bm{T}\bm{G}\right) = \tilde{\bm{W}}\tilde{\bm{G}}$, as in Eq.~(\ref{Eq: inllustration of uniqueness of lbaemaplsa}). From Theorem~\ref{theorem: lbaplsanmfsolution}, both $(\bm{M}, \bm{H})$ and $(\tilde{\bm{M}}, \tilde{\bm{H}})$ are solutions of NMF of $\bm{\Phi}$, where $\bm{M} = \bm{D}_{\bm{\Phi}}\bm{W}$, $\bm{H} = \bm{G}$, $\tilde{\bm{M}} = \bm{D}_{\bm{\Phi}}\bm{W}\bm{T}^{-1}$, $\tilde{\bm{H}} = \bm{T}\bm{G}$. Because $\bm{T}$ is not a permutation matrix, the solution of NMF is not unique.
    \end{proof}

A statement is true if and only if its contrapositive is true. Thus, from Theorem~\ref{theorem: mainresultsnonunique}, we immediately have:

\begin{theorem}\label{theorem: mainresultsunique}
The solution of NMF is unique if and only if the solution of LBA, EMA, and asymmetric PLSA is unique.
\end{theorem}

\subsection{Conclusion}\label{Sub: condis}

In this section, we review the identifiability issues of LBA, EMA, PLSA, and NMF. The idea of identifying solutions by making basis vectors as similar as possible --referred to as inner extreme solution, minimum end members, and minimum volume in LBA, EMA, and NMF, respectively -- is conceptually related across these models. However, the criteria used to quantify “as similar as possible” differ. In particular, the well-defined volume measure $\det(\bm{H}^T\bm{H})$ in NMF may partly explain why identifiability results in NMF are comparatively more developed. In addition, the assumption that basis vectors are as different as possible has been considered among LBA, EMA, and NMF.

We showed that the uniqueness of the NMF solution is closely related to the uniqueness of the solutions of LBA, EMA, and asymmetric PLSA. As a consequence, the well-developed uniqueness theory for NMF under assumptions such as separability, minimum volume, sparsity, and orthogonality can be transferred to LBA, EMA, asymmetric PLSA, LCA for two-way contingency tables, and symmetric PLSA. Conversely, the unique theorems related to the outer extreme solution and outermost end members in LBA and EMA are also suitable to NMF.

\section{Algorithms}\label{S: Algorithms/Estimators}

In NMF, the observed matrix $\bm{X}$ is partitioned in a signal part $\bm{\Phi} = \bm{MH}$ and an error part $\bm{E}_{\bm{X}}$ as $\bm{X} = \bm{MH} + \bm{E}_{\bm{X}}$.
Similarly, in LBA, EMA, and asymmetric PLSA, the observed conditional proportion $\bm{P}$ is partitioned in the signal part $\bm{\Pi} = \bm{WG}$ and the error part $\bm{E}_{\bm{P}}$ as $\bm{P} = \bm{WG} + \bm{E}_{\bm{P}}$. The objective function is closely related to the distributional assumptions of the data (e.g., the Frobenius norm for Gaussian data, the $l_1$ norm for Laplace data) \citep{van1992constrained, gillis2020nonnegative, SaberiMovahed2025Nonnegative}. Also, a regularizer can be added to the objective function, such as the determinant of $\bm{H}\bm{H}^T$ or the sum of distances between basis vectors \citep{ang2019algorithms, ZHANG2020106656, ang2025sum}. 

The objective function involving both the coefficient matrix and basis matrix is generally non-convex \citep{gillis2020nonnegative}. Thus, algorithms generally seek local optimal via alternatingly updating the basis and coefficient matrices. Some algorithms are one-stage, in which a final solution for the basis and coefficient matrices is obtained directly from $\bm{X}$ or $\bm{P}$ \citep{Leplat2019rank, ZHANG2020106656}. Some algorithms are two-stage, where unidentified estimates of basis and coefficient matrices are first computed and subsequently identified via the transformation matrix $\bm{S}$ in \eqref{Eq:illustrationofuniquenessofnmf} or $\bm{T}$ in \eqref{Eq: inllustration of uniqueness of lbaemaplsa}
\citep{weltje1997end, mooijaart1999least, fu2018identifiability}.

\subsection{LBA}

LBA estimation, where the model is identified using the inner extreme solution or outer extreme solution as mentioned in Section~\ref{Sub: identlba}, is typically two-stage as described above. In the first stage, for frequency data, one may assume that the observations follow a product-multinomial distribution; 
a maximum likelihood estimator (MLE) is computed by employing the idea of majorization from convex analysis \citep{de1990latent, Sun2017Majorization} or equivalently the expectation maximization
(EM) algorithm \citep{Dempster1977Maximum, van1992constrained}. Alternatively, a constrained weighted least squares estimator, with no distributional assumptions, is computed by the active
constraint method \citep{mooijaart1999least}. In the second stage, the transformation matrix $\bm{T}$ by the Metropolis algorithm is determined to minimize or maximize the sum of chi-squares distances between the rows of basis vectors $\bm{TG}$ (See Eq.~(\ref{E: lbamodchisquareinnerouter})), yielding inner or outer extreme solutions. The R package $lba$ implements these methods \citep{RJ-2018-026}.

\subsection{Symmetric and Asymmetric PLSA}

PLSA is based on the statistical model LCA and uses the EM algorithm for MLE \citep{goodman1974exploratory, goodman1987new, hofmann1999learning, hofmann1999probabilistic, hofmann2001unsupervised}. To mitigate overfitting, a temperature parameter $\beta \in (0, 1]$ is introduced
to the EM algorithm. At $\beta = 1$, the algorithm reduces to the standard EM algorithm. In this case, the EM estimates of the asymmetric PLSA are identical to the EM estimates obtained from LBA, whereas the EM estimates of the symmetric PLSA are identical to those obtained from LCA. The R package $svs$ provides implementations \citep{Packagesvs2024}.

\subsection{EMA}

The end-member modeling algorithm (EMMA), proposed by \citet{weltje1997end}, has been a benchmark algorithm for finding minimum end members introduced in Section~\ref{Sub: identlba} \citep{weltje2007genetically, van2018genetically, Dietze2022application}. EMMA first obtains a nonnegative lower-rank approximation via SVD with nonnegativity corrections, then applies a simplex expansion algorithm to find minimum end members. An alternative EMA algorithm is AnalySize based on HALS-NMF instead of SVD; NMF is an appealing approach for EMA, eliminating the need for nonnegativity corrections  \citep{Chen2012HALSNMF, paterson2015new, qi2026madnmf}. For a highly mixed dataset where no single data point is near a basis vector, BasEMMA seeks the outermost basis vectors based on a genetic algorithm \citep{ZHANG2020106656}.
Alternatively, \citet{qi2026unmixing} proposed an alternating projected fast gradient method (APFGM) for finding outermost end members. \citet{seidel2015r} created R functions for EMMA proposed by \citet{weltje1997end}.  \citet{paterson2015new} provides AnalySize based on MATLAB. BasEMMA is embedded in the Microsoft Excel program \citep{ZHANG2020106656}, and APFGM is performed in MATLAB \citep{qi2026unmixing}. In addition, there are other algorithms and implementations such as EMMAgeo \citep{DIETZE2012169, egqsj-68-29-2019} and BEMMA \citep{Yu2016}.

\subsection{NMF}

Since \citet{lee1999learning}'s publication in $Nature$, many algorithms have been developed for NMF, see \citet{Fu2019Nonnegative, ang2019algorithms, gillis2020nonnegative, GUO2024102379, SaberiMovahed2025Nonnegative} for surveys. As discussed in Section~\ref{Sub: sufnmf}, under separability, basis vectors are assumed to be present in the data; algorithms such as VCA, SPA, SNPA select $K$ rows of $\bm{X}$ to form the basis matrix and then the coefficient matrix can be estimated by solving a constrained least squares problem \citep{nascimento2005vca, araujo2001successive, GillisSNPA2014,
barbarino2025robustness}. Under the minimum-volume assumption, some methods minimize a regularized objective consisting of a data-fitting term and a volume regularizer (e.g., \citet{2007ITGRS45765M, Zhou6003792, fu2016robust, ang2019algorithms, Leplat2019rank, leplat2020}); some methods minimize the volume of the basis matrix subject to the constraint $\bm{X} = \bm{MH}$, including MVSA \citep{Li2008}, SISAL \citep{Bioucas5289072}, and MVES \citep{Chan2009Convex}. Optimization techniques used in these methods include projected gradient descent, linear programming, quadratic programming, multiplicative updates, augmented Lagrangian, among others.
Methods beyond separability and minimum volume assumptions have also been proposed, including \citet{Cai2011graph, ge2015intersecting, Lin2018maximum, javadi2020nonnegative, abdolali2021simplex, Abdolali2024Dual, ang2025sum, qi2026identification, qi2026madnmf}. Code implementations are widely available
in MATLAB, Python, and R \citep{GUO2024102379}.

\section{A geological example: sediment transport in the Slufter tidal-inlet system
}\label{S: geologyexampleslufter}

\subsection{Introduction}

\begin{figure}[h]
\centering
  \begin{subfigure}[b]{0.38\linewidth}
\includegraphics[width=1\textwidth]{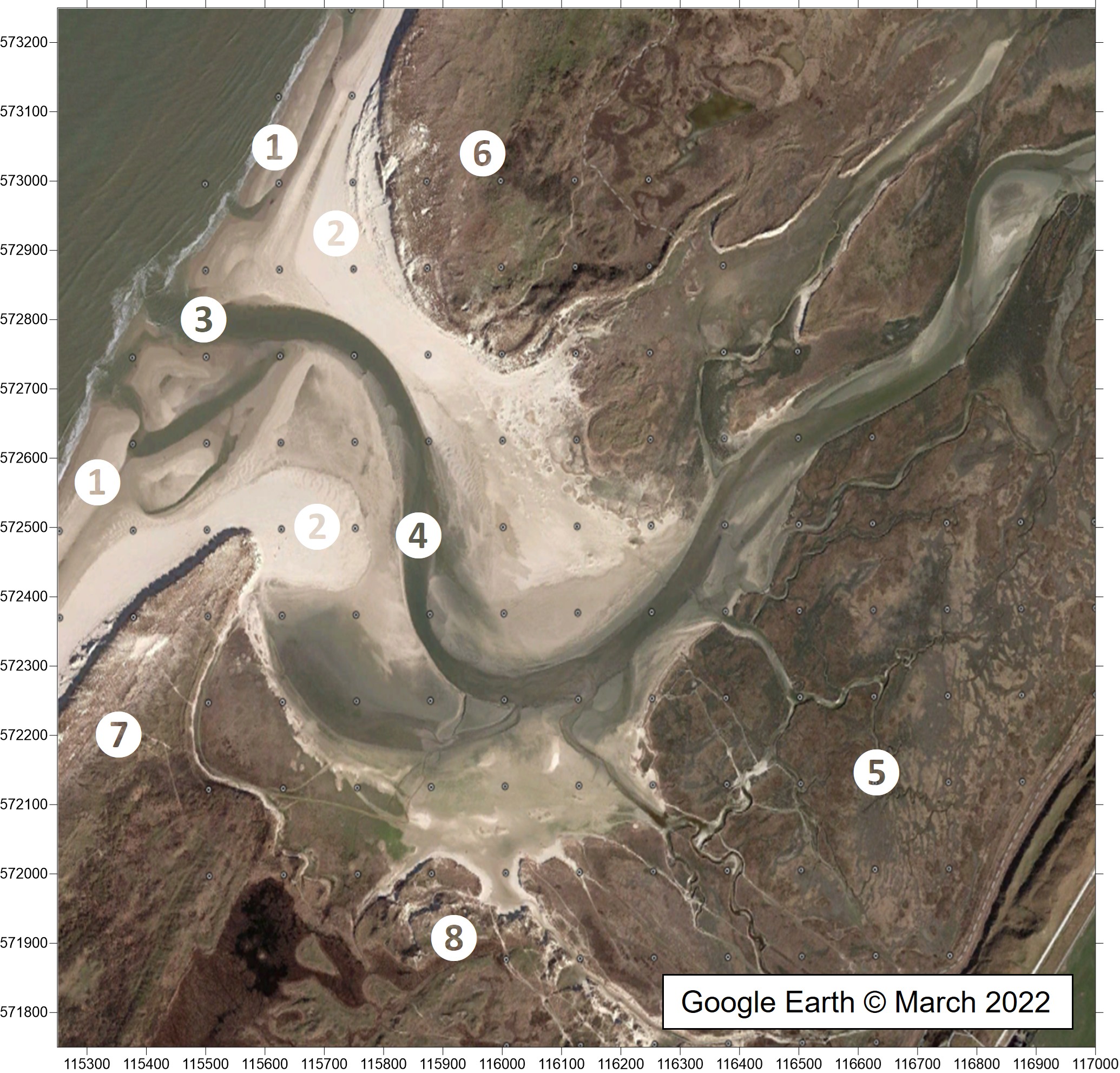}
 \caption{\scriptsize{Satellite image}}\label{F: gsdF1a}
 \end{subfigure}
   \begin{subfigure}[b]{0.38\linewidth}
\includegraphics[width=1\textwidth]{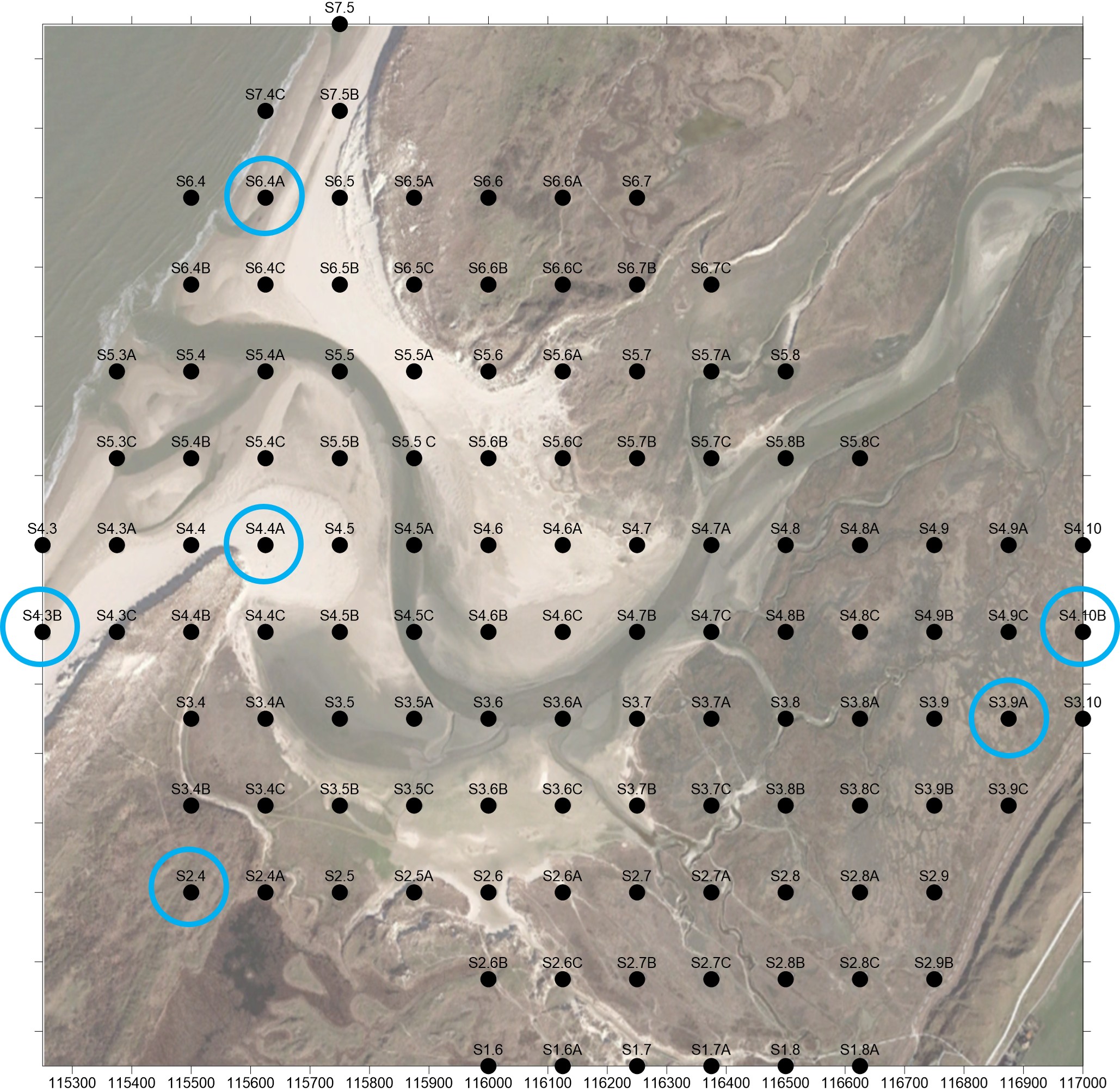}
    \caption{\tiny{Sampling grid composed of 118 locations}}
    \label{F: gsdF1b}
 \end{subfigure}
 \caption{(a) Satellite image (Google Earth, March 2022) showing the southern part of the tidal inlet system the Slufter, Texel, the Netherlands. Numbers indicate the location of: (1) ‘wet’ beach, (2) ‘dry’ beach, (3) mouth of the main tidal channel, (4) inner parts of the main channel, (5) salt marsh, (6) foredune ridge (sand-drift dike) north and (7) south of the tidal channel, (8) the Slufterbollen coastal dunes. (b) Sampling grid composed of 118 locations where surface sediment samples have been obtained (March/April 2021). Field impressions of the samples highlighted with the blue circles are illustrated in Figure~\ref{F: gsdF2}; those samples are selected as they are dominated by one of the modelled end members of the 6-end-member model.}\label{F: gsdF1}
    \end{figure}

Here we present a geological application of the end-member modelling approach for determination of sediment transport pathways based on end-member modelling of grain size distribution data from laser-diffraction particle size analysis. Surface sediment samples from an active secondary tidal inlet system along the Dutch coast - the Slufter nature reserve on the island of Texel (Figure~\ref{F: gsdF1a}) - are used to assess the physical meaning of the method’s output in a complex sedimentary system where sediment is transported by waves, tides and wind. The Slufter is a unique, dynamic secondary tidal inlet system that allows the North Sea to flood a low-lying dune valley through a narrow tidal channel. Sediment transport within the tidal inlet system is dominated by the flood and ebb tidal currents, both transporting sand and mud, and thereby creating a saltwater marsh habitat in the sheltered eastern part of the system. The Slufter system includes sandy beaches in the west, where sediment transport is dominated by waves (‘wet’ beach) and wind (‘dry’ beach). Sediment transport in the foredune ridge system (sand-drift dike) is dominated by aeolian (wind) transport supplying sandy sediments from the beach to the coastal dunes. The grain size distribution dataset is decomposed into a series of primary sedimentary components by end-member modelling. The spatial distribution of these components is controlled by the dominant transporting media (sea water, wind), the mode of transport (creep, saltation, suspension), geomorphology and vegetation cover and reflects the main processes of sediment transport in this mixed aeolian-tidal sedimentary system.  

\subsection{Material and methods}

\begin{figure}[h]
\centering
  \begin{subfigure}[b]{0.38\linewidth}
\includegraphics[width=1\textwidth]{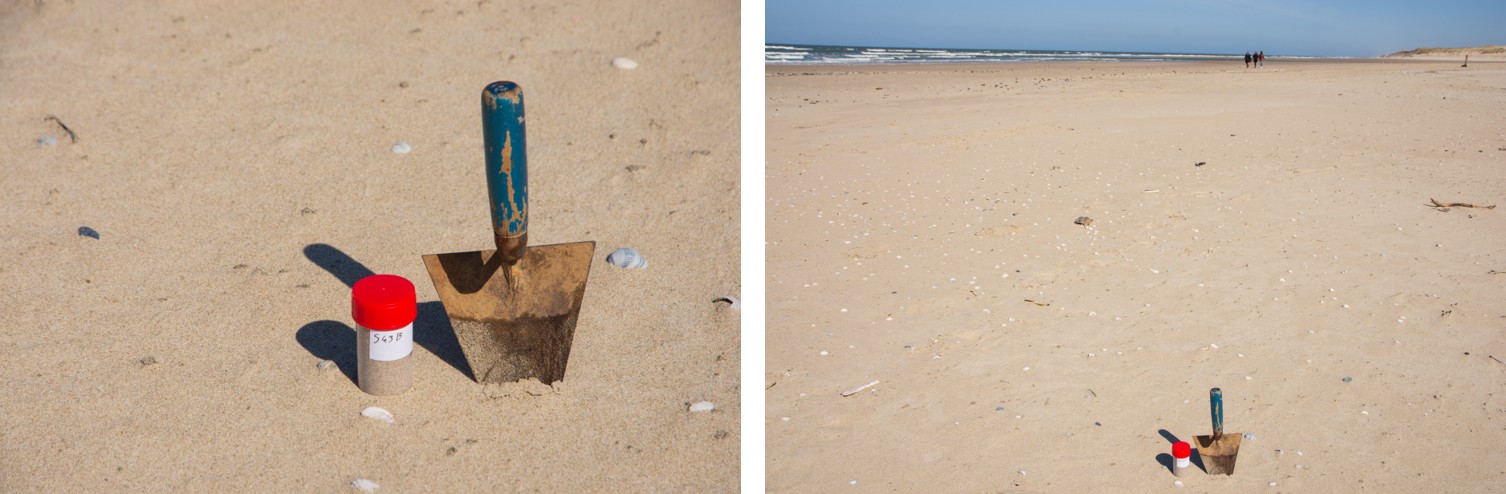}
 \caption{Sample S4.3B}\label{F: gsdF2a}
 \end{subfigure}
 \hspace{0.5cm}
   \begin{subfigure}[b]{0.38\linewidth}
\includegraphics[width=1\textwidth]{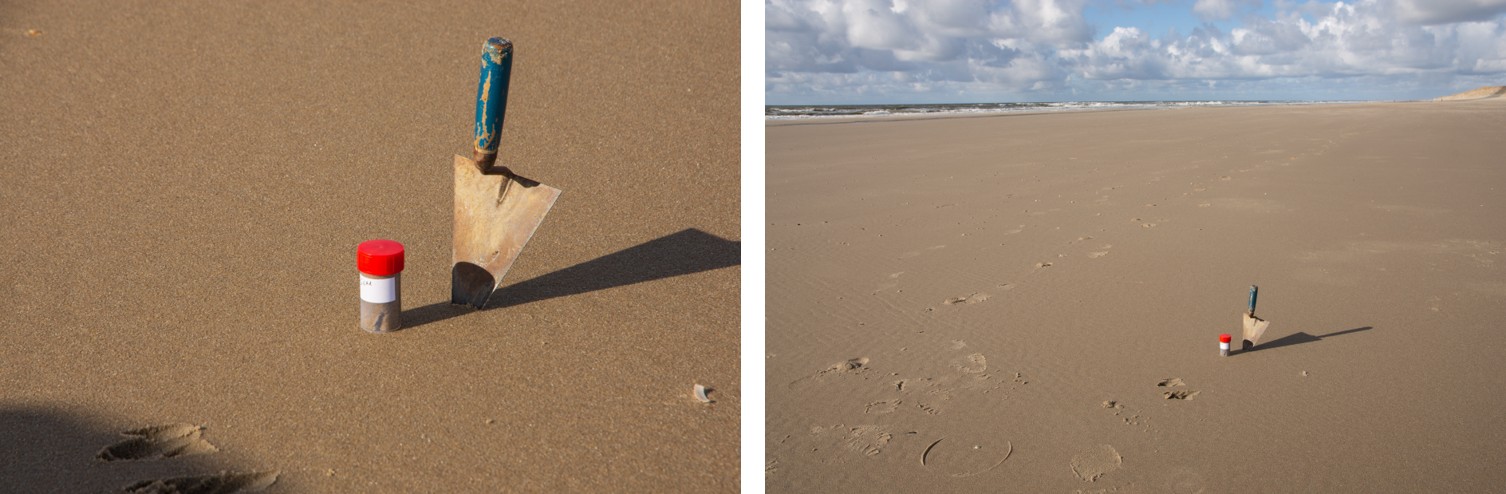}
    \caption{Sample S6.4A}
    \label{F: gsdF2b}
 \end{subfigure}
 \hspace{0.5cm}
   \begin{subfigure}[b]{0.38\linewidth}
\includegraphics[width=1\textwidth]{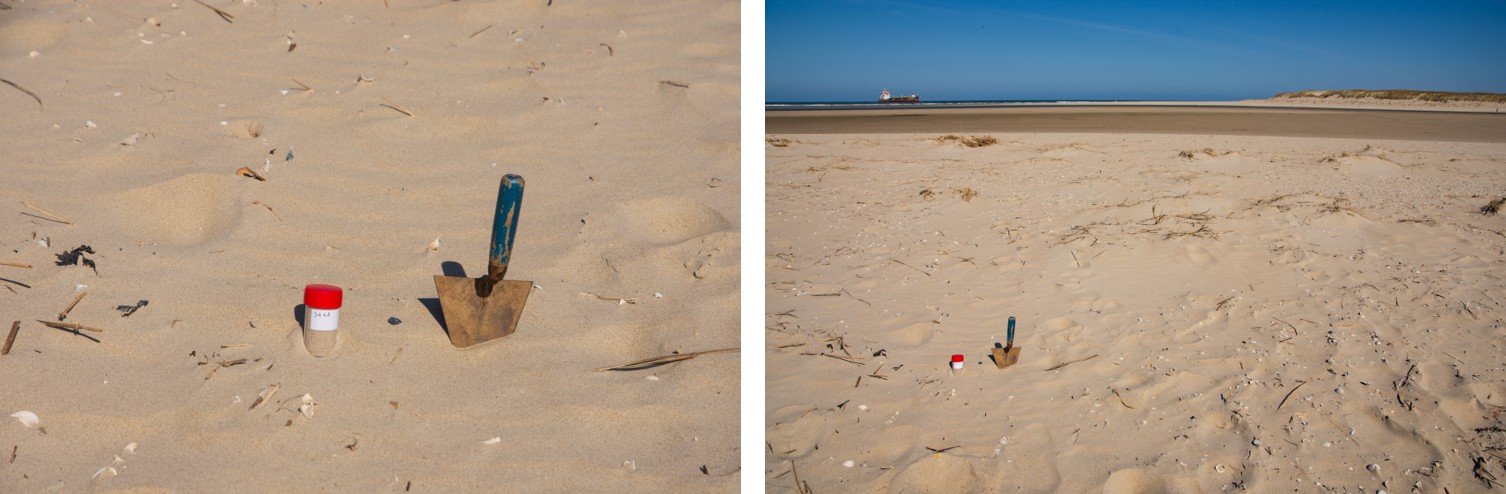}
 \caption{Sample S4.4A}\label{F: gsdF2c}
 \end{subfigure}
 \hspace{0.5cm}
   \begin{subfigure}[b]{0.38\linewidth}
\includegraphics[width=1\textwidth]{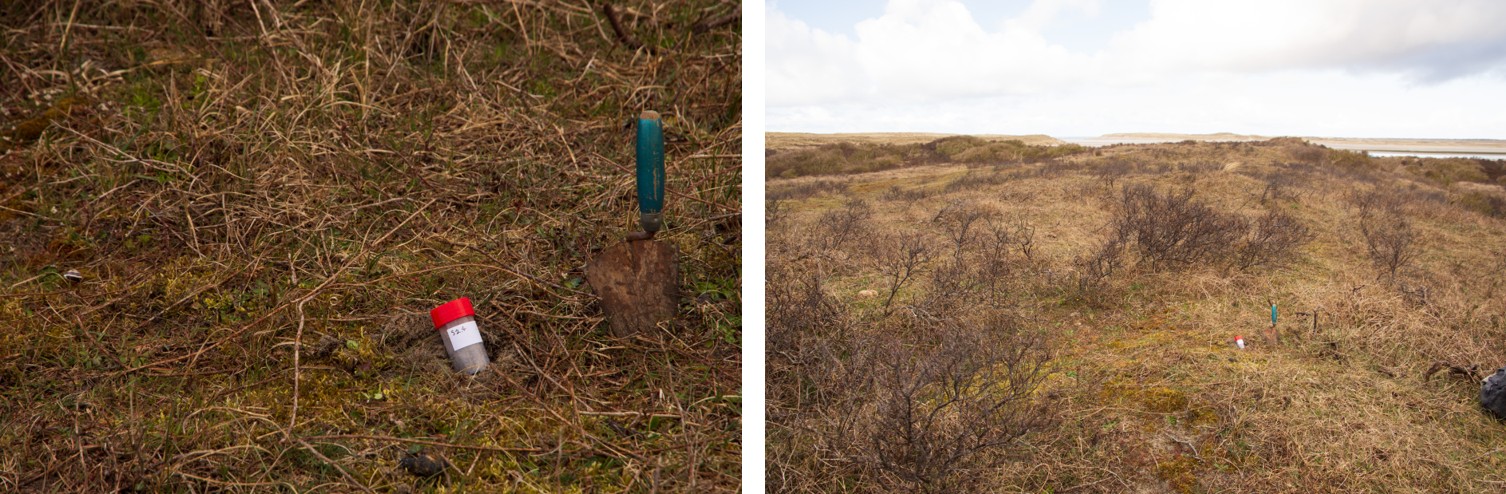}
    \caption{Sample S2.4}
    \label{F: gsdF2d}
 \end{subfigure}
 \hspace{0.5cm}
   \begin{subfigure}[b]{0.38\linewidth}
\includegraphics[width=1\textwidth]{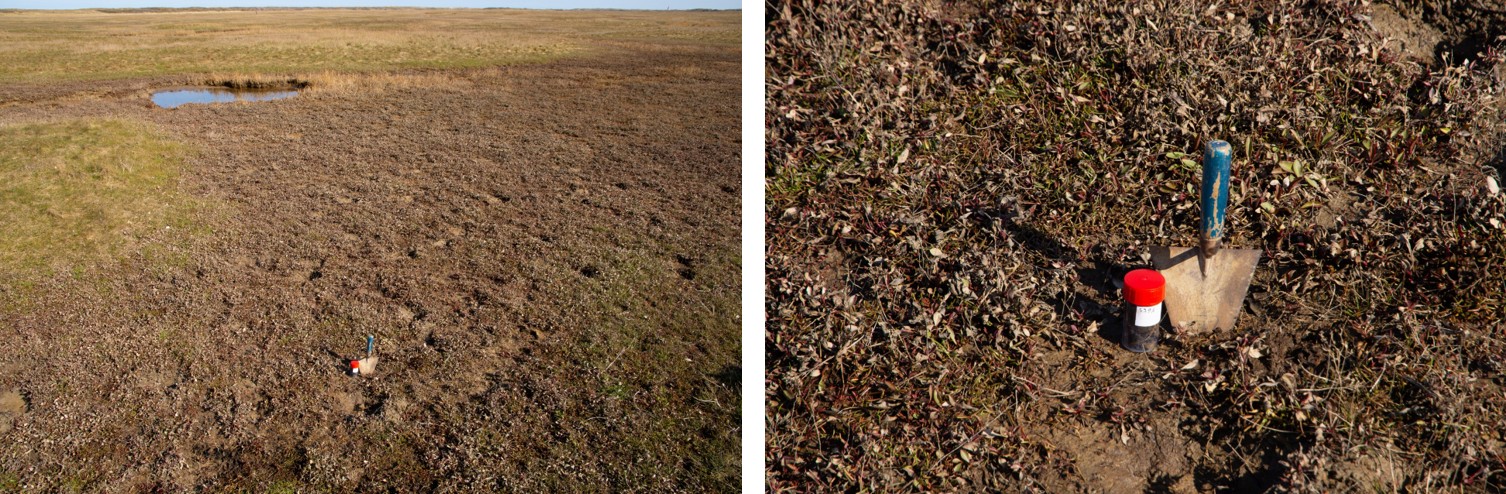}
 \caption{Sample S3.9A}\label{F: gsdF2e}
 \end{subfigure}
 \hspace{0.5cm}
   \begin{subfigure}[b]{0.38\linewidth}
\includegraphics[width=1\textwidth]{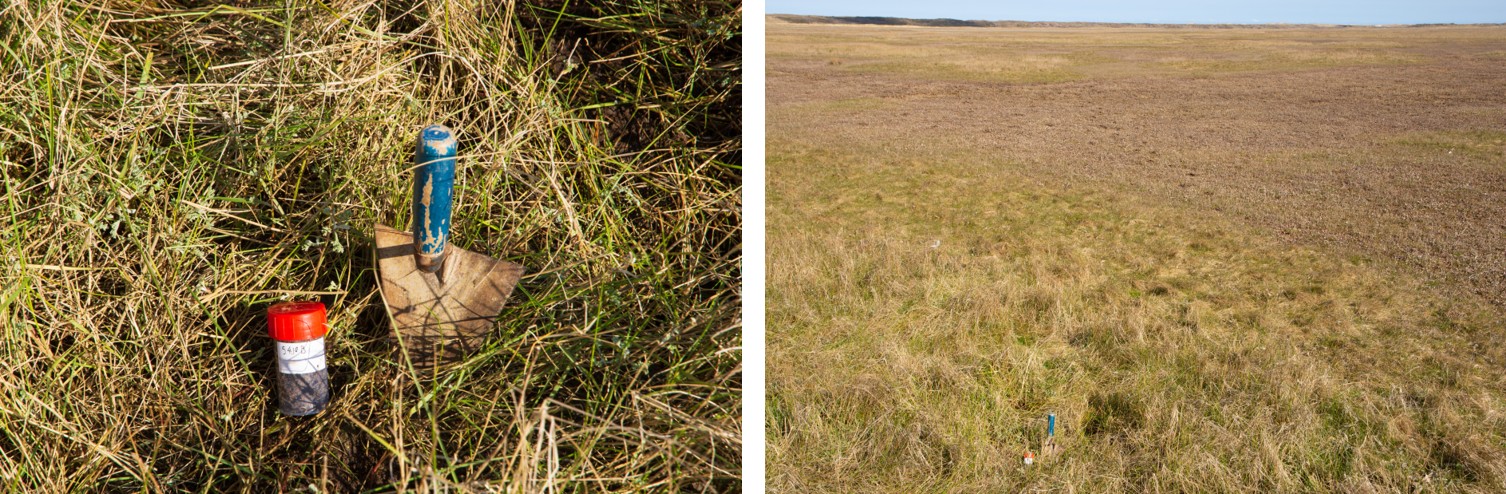}
    \caption{Sample S4.10B}
    \label{F: gsdF2f}
 \end{subfigure}
 \caption{Field impressions of six surface samples (for exact locations see Figure~\ref{F: gsdF1b}) which are selected as they are dominated by one of the modelled end members of the 6-end-member model. (a) sample S4.3B from the high water mark, boundary between the ‘wet’ and ‘dry’ beach (EM1-dominated), (b) sample S6.4A from the ‘wet’ beach (EM2-dominated), (c) sample S4.4A from the ‘dry’ beach (EM3-dominated), (d) sample S2.4 from the densely vegetated foredune ridge (EM4-dominated), (e) sample S3.9A (EM5-dominated) and (f) sample S4.10B (EM6-dominated), both from the densely vegetated salt marsh.}\label{F: gsdF2}
    \end{figure}

Surface sediment samples ($I=118$) have been obtained on a fixed grid with a spatial resolution of 125 m in March-April 2021 (Figure~\ref{F: gsdF1b} and Figure~\ref{F: gsdF2}). A Sympatec HELOS KR laser particle sizer was used to analyse the grain-size distributions (57 size classes in the range 0.12-2000 µm) after chemically removing organic matter (plant remains) and carbonates (e.g. shell fragments).

The row totals of the grain-size distribution dataset did not sum to 1, owing to rounding errors. The row totals are made constant. Thus, NMF, asymmetric PLSA, LBA, and EMA are equivalent if we constrain the basis matrix and coefficient matrix in NMF as in LBA, EMA, and asymmetric PLSA. 

The grain-size distributions of all samples are decomposed using end-member modelling in various end-member mixing models with the number of end members varying between 2 and 8. The first modelling stage involves the estimation of the number of end members. The second modelling stage involves estimation of the end member compositions, and subsequently, calculation of the proportional contribution of the end members in all observations. We employ algorithms developed in the context of NMF.

\subsection{Results and interpretation}

\subsubsection{Estimation of the number of end members}

\begin{figure}
\centering
\begin{subfigure}[b]{0.38\linewidth}
\includegraphics[width=1\textwidth]{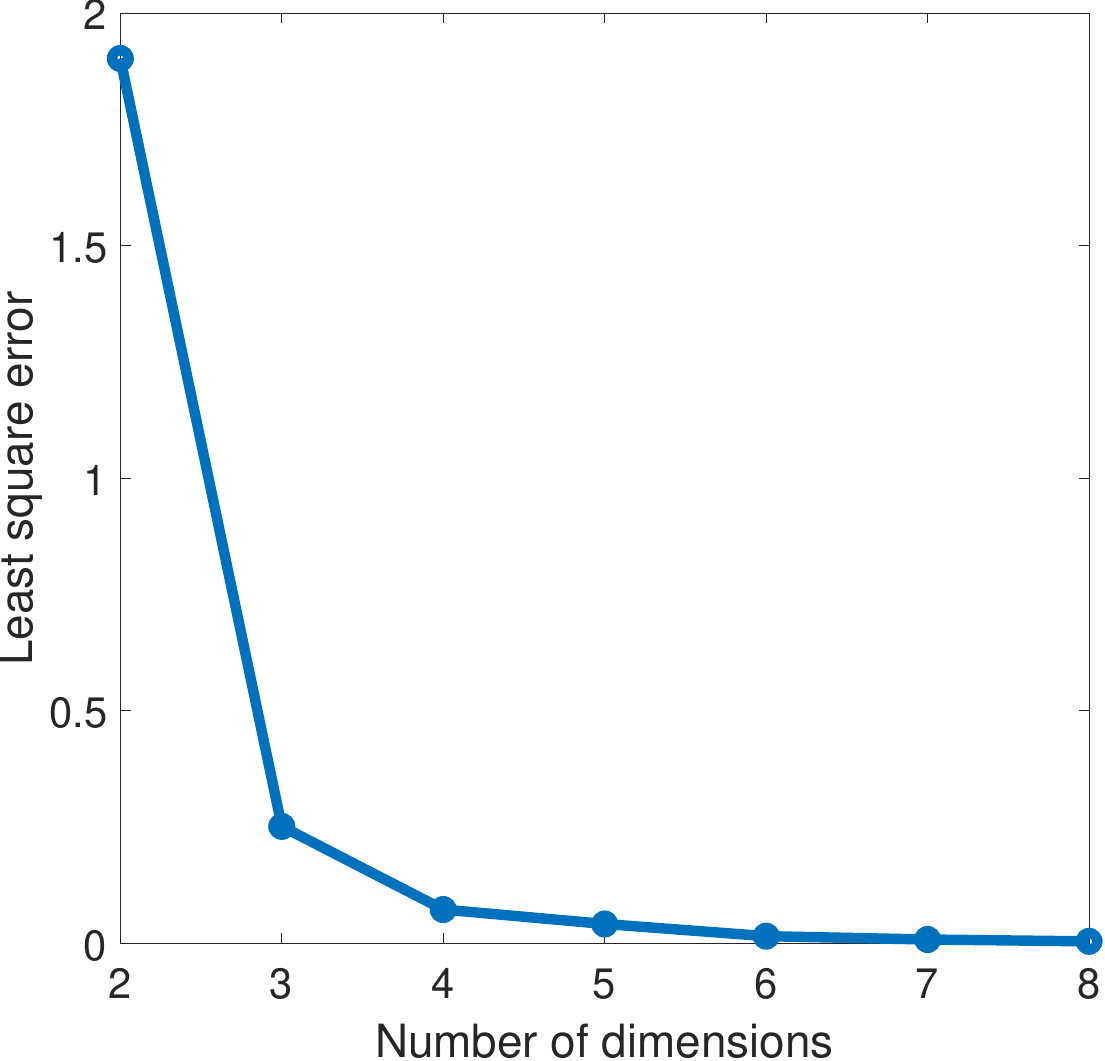}
  \caption{Least square error}
 \label{F: sgd2021minvolleastsquareerror}
 \end{subfigure}
    \begin{subfigure}[b]{0.435\linewidth}
\includegraphics[width=1\textwidth]{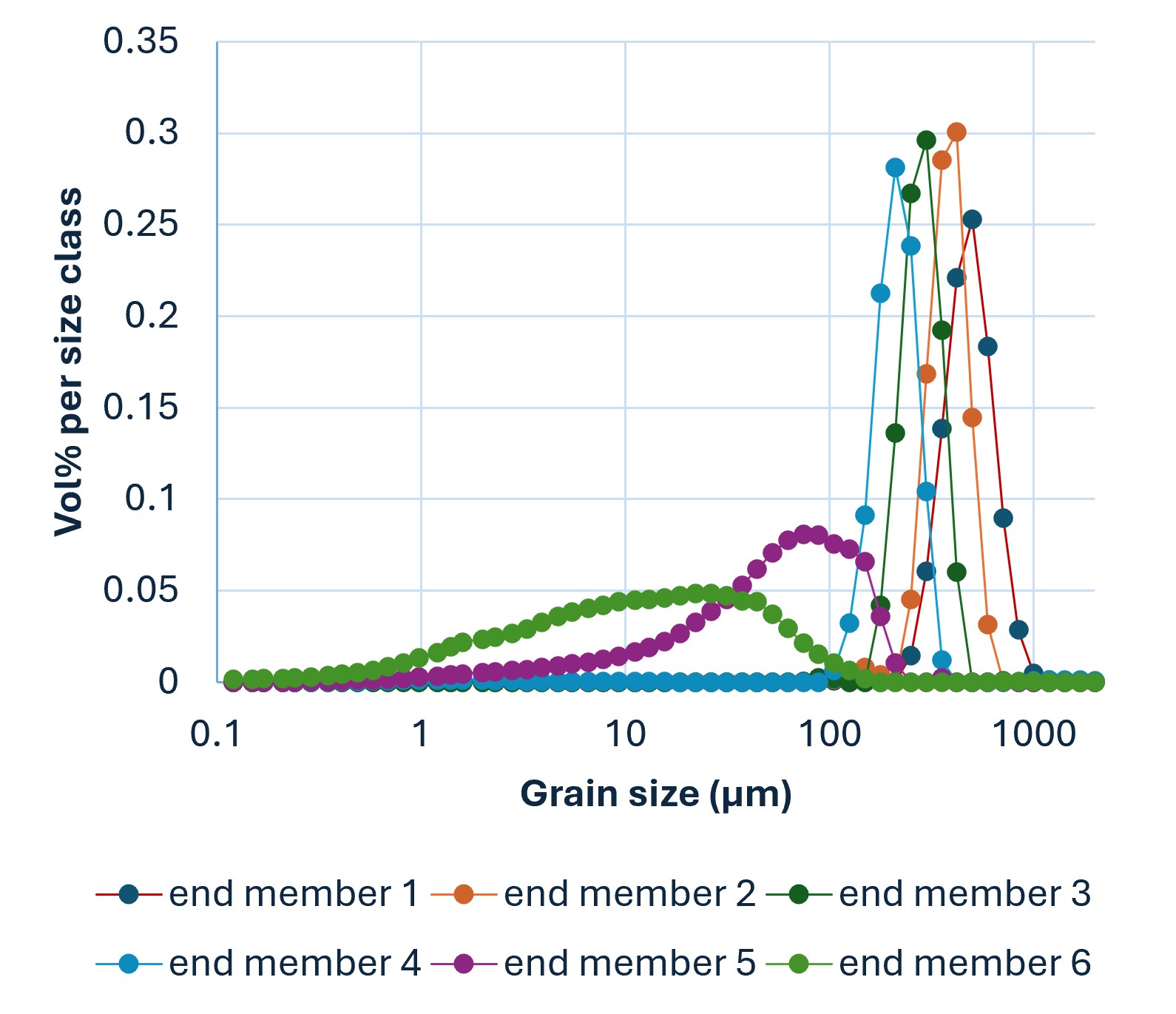}
 \caption{Six end members}\label{F: gsdF3b}
 \end{subfigure}
 \caption{(a) Least square error; (b) Grain-size distributions of the modelled end members of the 6-end-member model. Labelling of the end members follows the modal size: EM1 = coarsest-grained end member, EM6 = finest-grained end member. Modal size of the end members: EM1 500 µm, EM2 420 µm, EM3 297 µm, EM4 210 µm, EM5 74 µm and EM6 26 µm.}\label{F: sgd2021minvolchoosedimdim6}
    \end{figure}

Figure~\ref{F: sgd2021minvolleastsquareerror} shows the error from NMF for the number of dimensions ranging from 2 to 8. From the plot we can see that, when the number of dimensions is 6, the observed data are approximated to a high degree by the basis vectors and coefficient matrix. Therefore, we choose the 6-end-member model.

\subsubsection{Six end member model}
The grain-size distributions of the six modelled end members are shown in Figure~\ref{F: gsdF3b}. All six end members are characterised by unimodal, partly overlapping distributions with clearly defined modes at 500 µm (EM1), 420 µm (EM2), 297 µm (EM3), 210 µm (EM4), 74 µm (EM5) and 26 µm (EM6). The size range of EM1 to EM4 fall within the sand range (by definition 63-2000 µm), whereas EM5 is best typified as sandy silt (with silt between 8-63 µm) and EM6 as clayey silt (with clay $<$8µm). The sandy end members (EM1-EM4) are very well sorted, reflected by their narrowly distributed particle sizes, whereas the sandy silt (EM5) and clayey silt (EM6) end members are very broadly distributed.

\subsubsection{Spatial distribution maps of the modelled end members}

\begin{figure}
\centering
  \begin{subfigure}[b]{0.38\linewidth}
\includegraphics[width=1\textwidth]{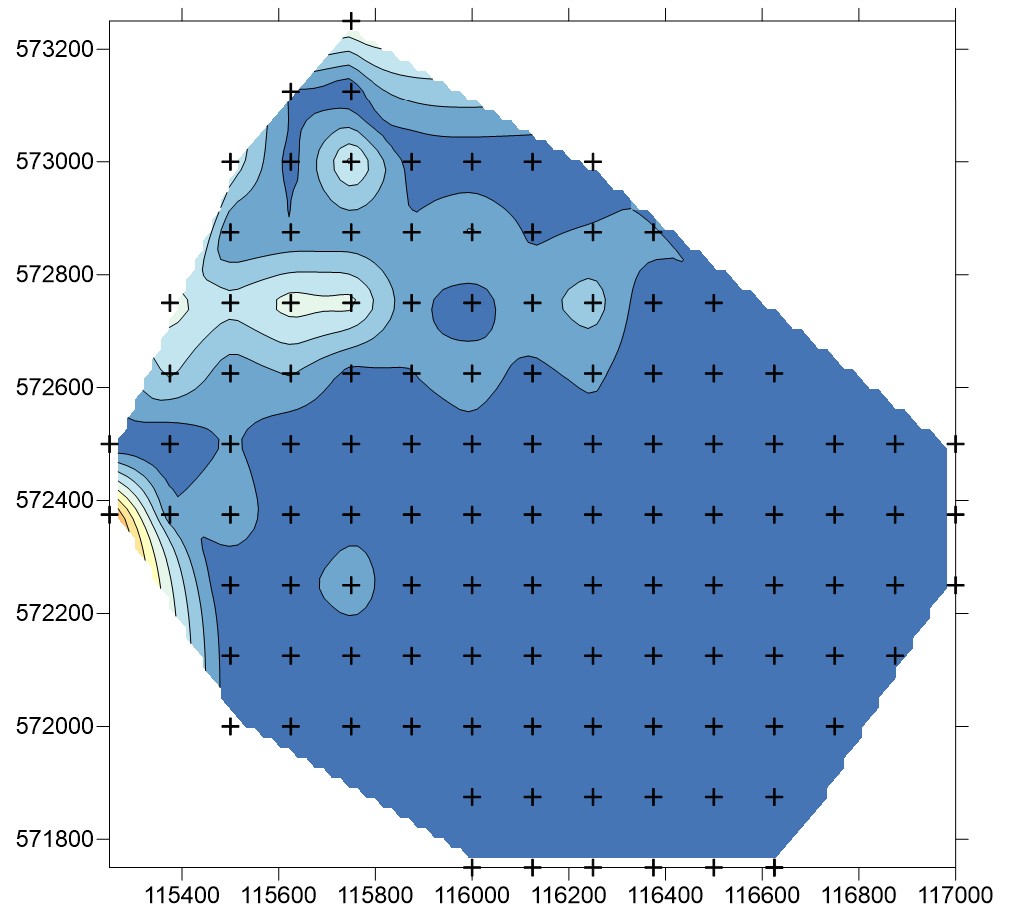}
 \caption{EM1}\label{F: gsdF4a}
 \end{subfigure}
   \begin{subfigure}[b]{0.38\linewidth}
\includegraphics[width=1\textwidth]{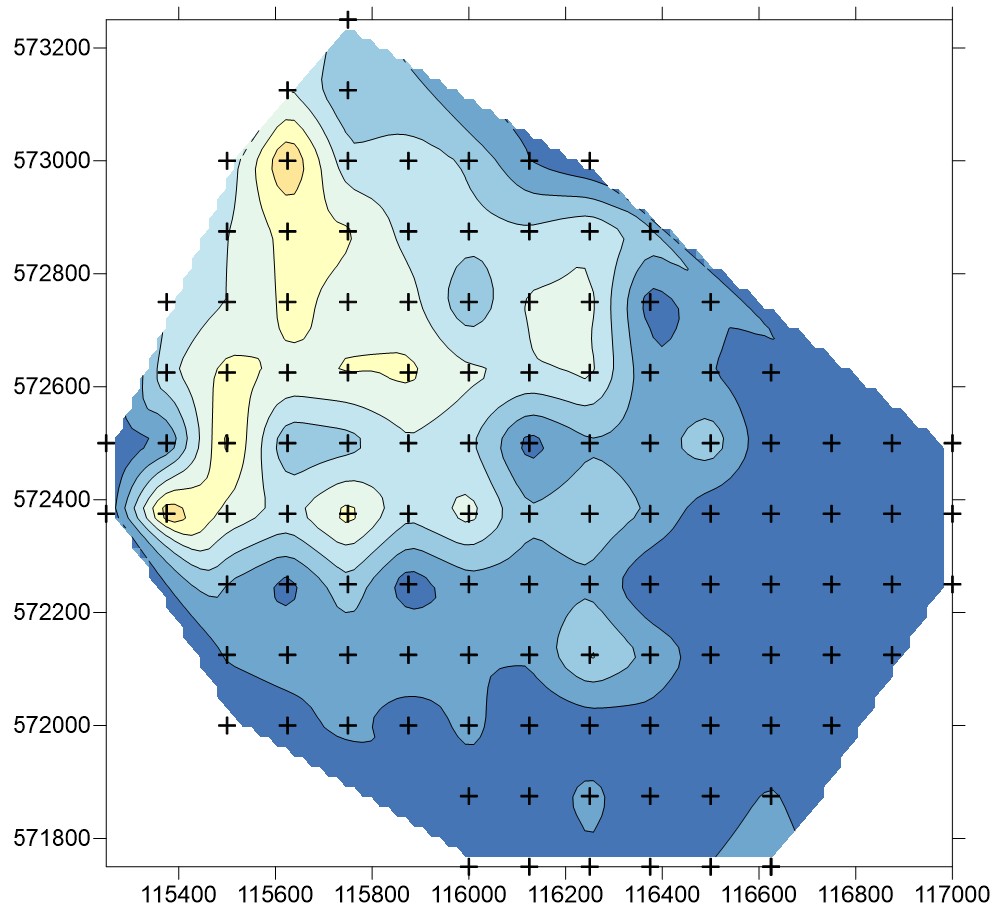}
    \caption{EM2}
    \label{F: gsdF4b}
 \end{subfigure}
   \begin{subfigure}[b]{0.38\linewidth}
\includegraphics[width=1\textwidth]{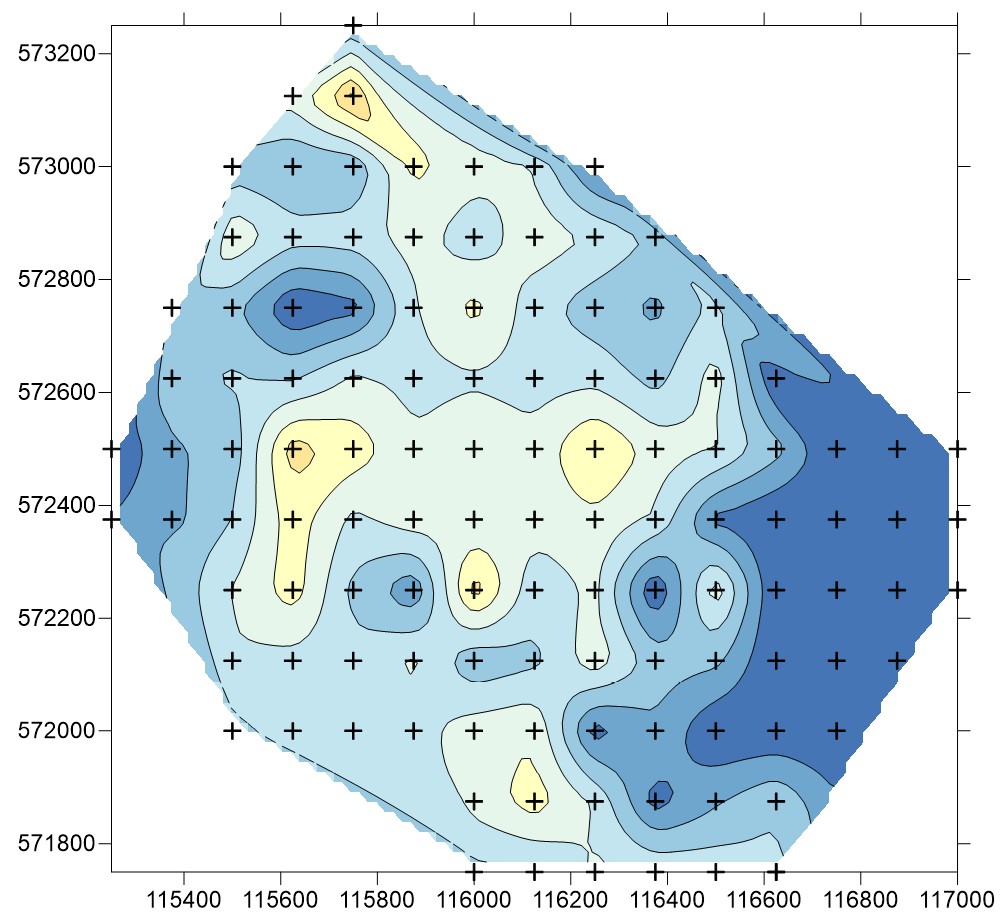}
 \caption{EM3}\label{F: gsdF4c}
 \end{subfigure}
   \begin{subfigure}[b]{0.38\linewidth}
\includegraphics[width=1\textwidth]{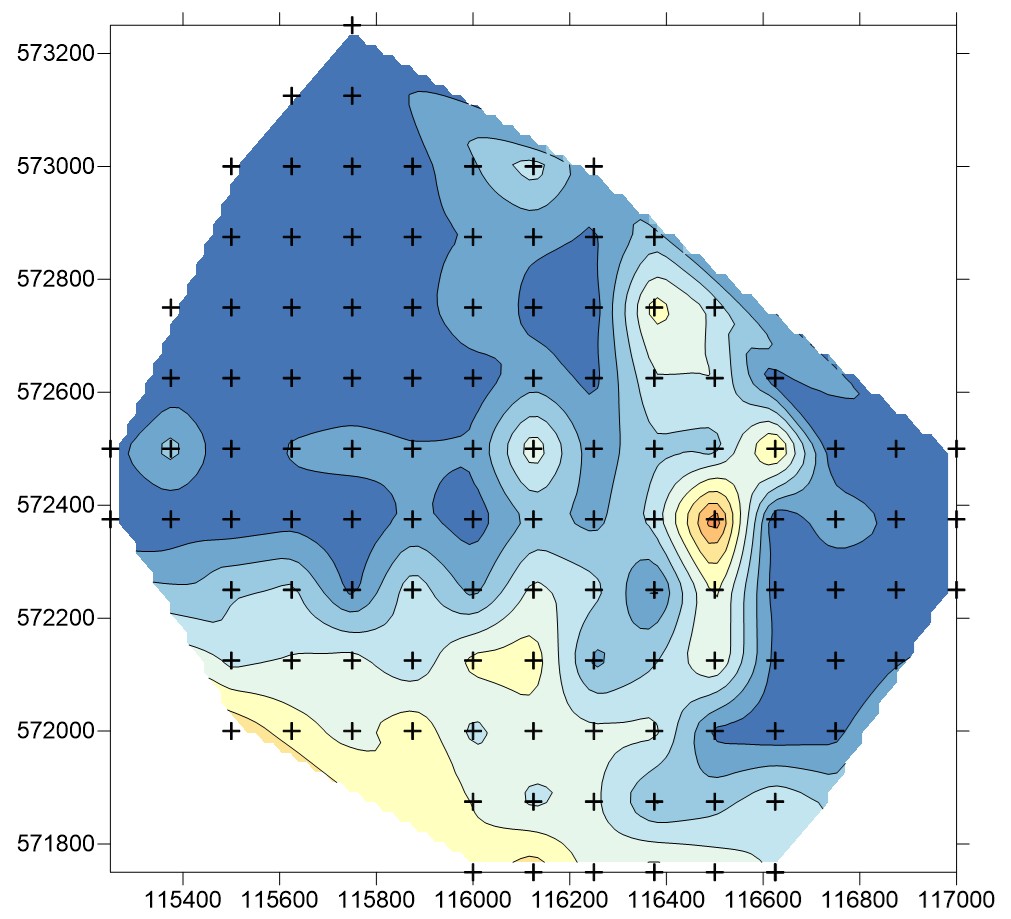}
    \caption{EM4}
    \label{F: gsdF4d}
 \end{subfigure}
   \begin{subfigure}[b]{0.38\linewidth}
\includegraphics[width=1\textwidth]{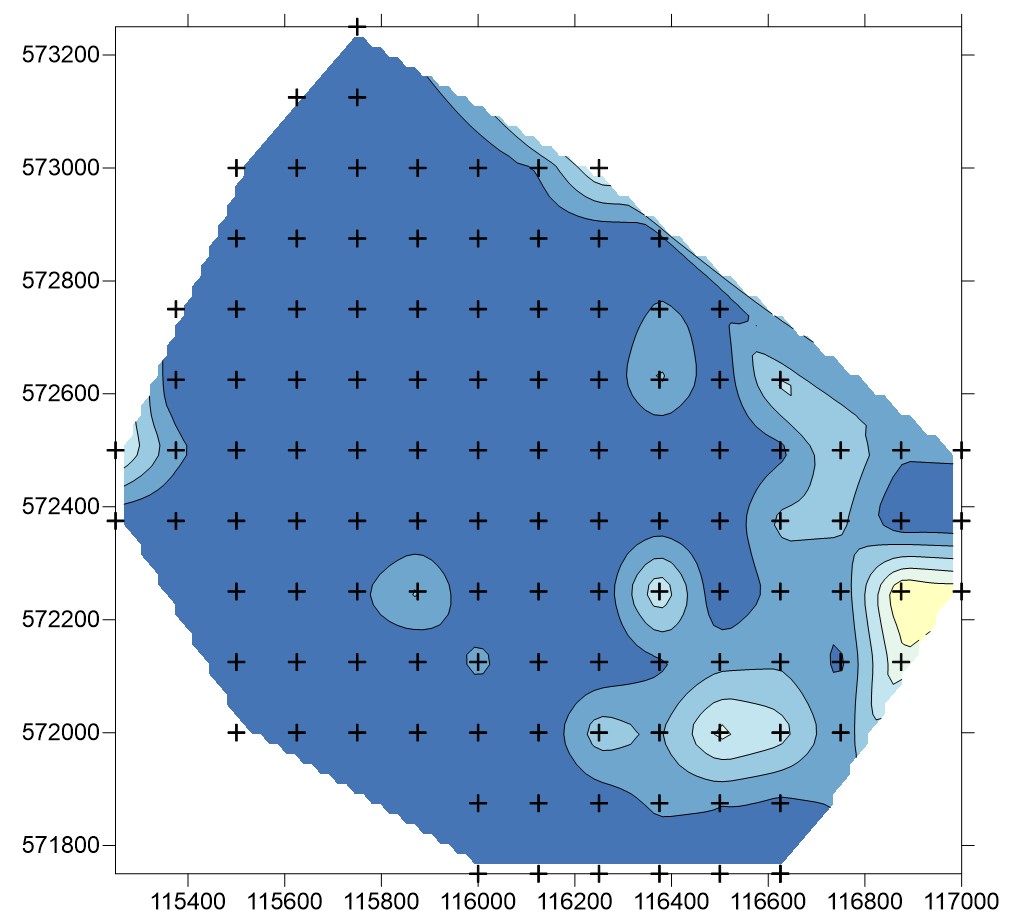}
 \caption{EM5}\label{F: gsdF4e}
 \end{subfigure}
   \begin{subfigure}[b]{0.38\linewidth}
\includegraphics[width=1\textwidth]{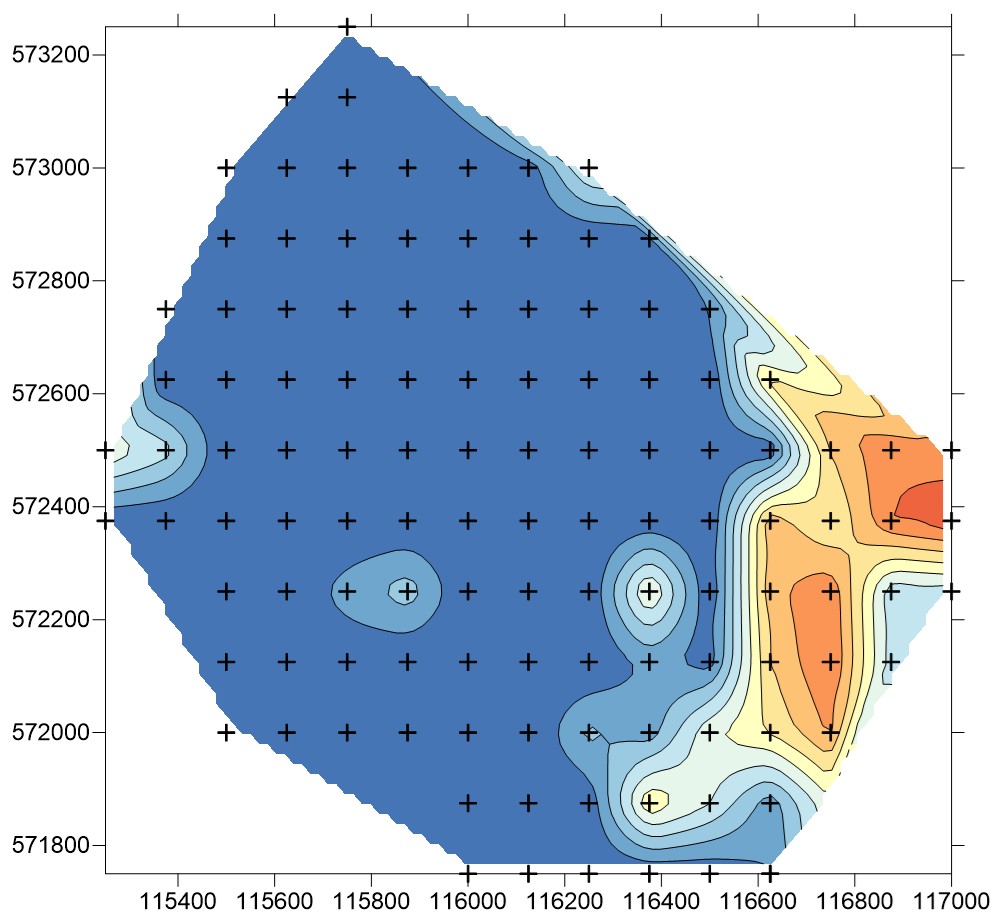}
    \caption{EM6}
    \label{F: gsdF4f}
 \end{subfigure}
 \vspace{0.5em}
\begin{center}
\includegraphics[angle=90, width=0.6\linewidth]{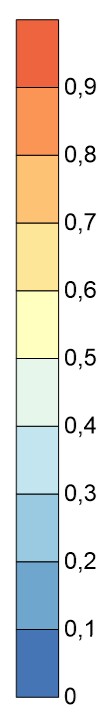}
\end{center}
 \caption{Spatial distribution maps of the proportional abundances of the modelled end members of the 6-end-member model. (a) end member EM1, (b) EM2, (c) EM3, (d) EM4, (e) EM5 and (f) EM6.}\label{F: gsdF4}
    \end{figure}

Each of the spatial end member distribution maps (Figure~\ref{F: gsdF4}) show some distinct regions where each of the end members dominates the well-mixed sediment composition.

End member EM1 shows elevated proportional abundances near the high-water line on the beach and the mouth of the main tidal channel (Figures~\ref{F: gsdF2a} and \ref{F: gsdF4a}), both high-energetic sub-environments, where bedload transport of coarse sand ($\sim$500 µm) occurs by waves and tidal currents. 

End member EM2 is abundant in a broad zone including large parts of the beach and the sand flats near the mouth and inner parts of the main channel (Figures~\ref{F: gsdF2b} and \ref{F: gsdF4b}). Overall, the EM2-dominated zone is located landward of the EM1-dominated zone, indicating that these finer-grained sands ($\sim$420 µm) are transported as bedload by slightly lower-energetic waves and tidal currents.

End member EM3 is especially abundant in the samples obtained from the ‘dry’ beach, the marram grass dominated foredune ridges located north and south of the main channel and the dry parts of the sand flats bordering the inner parts of the main channel (Figures~\ref{F: gsdF2c} and \ref{F: gsdF4c}). These locations are all dominated by aeolian bedload transport of relatively coarse-grained aeolian sand ($\sim$300 µm). Compared to the two coarser-grained end members (EM1, EM2), end member EM3 dominates the sediments located landward of the EM1- and EM2-dominated zones.

End member EM4 continues this spatial trends as it reaches its highest abundances even more landward, especially in the sheltered, densely vegetated inland dunes (Slufterbollen) and sand flats located south of the main channel, and in a narrow zone east of the main channel in the westernmost part of the salt marsh (Figures~\ref{F: gsdF2d} and \ref{F: gsdF4d}). These sub-environments are all dominated by aeolian bedload transport of relatively fine-grained aeolian sand ($\sim$210 µm).

End members EM5 and EM6 (Figures \ref{F: gsdF2e}-\ref{F: gsdF2f} and \ref{F: gsdF4e}-\ref{F: gsdF4f}) show very distinct zones of high abundances which are exclusively bounded to the salt marsh area, so dominating the eastern most part of the study area. The very fine-grained sandy and clayey silts ($\sim$26-74 µm) are transported into the Slufter system by the flood tidal currents as suspension load, and deposition of these fines is controlled by the salt marsh vegetation.

\subsection{Conclusions}

End-member modelling of the grain-size distribution dataset composed of 118 surface sediment samples from the Slufter tidal-inlet system allowed the distinction of six sedimentary components which are characterized by unimodal grain-size distribution with modal sizes varying between 26 and 500 µm. The sandy end members have distinctly different modal sizes but are all very well sorted during bedload transport under the influence of waves and tidal currents (EM1, EM2) and by wind (EM3, EM4). The finest-grained end members (EM5, EM6) have broadly distributed, poorly sorted grain-size distribution and are dominated by the mud fraction (clay, silt) transported as suspension load by flood tidal currents.

Geomorphology and vegetation cover control the influence of the transporting agent (water, wind) and mode of sediment transport and deposition (bedload, suspension load). The spatial distribution of the end members therefore reflect lateral differences in energetic conditions within the different sedimentary sub-environments, i.e. reflect differences in the intensity of transporting agents – waves, tidal currents, wind – that transport and deposit sediments, ranging from high-energy (rapid, turbulent) to low-energy (still, quiet) conditions. The west-to-east trends in sediment composition, e.g. the ratio sandy versus silty end members and the relative abundances of the sandy end member in the sand fraction, clearly indicates that energetic conditions gradually change from high-energy at the beach and near the mouth of the main tidal channel, to intermediate-energy along the inner parts of the tidal channel and in the dune areas, and finally towards low-energy at the salt marsh system.

\section{Related model: Archetypal analysis}\label{S: Related models/methods}

Archetypal analysis (AA) is proposed by \citet{cutler1994archetypal}. In AA, basis vectors $\bm{h}_1, \cdots, \bm{h}_K$ are convex combinations of data points $\bm{x}_1, \cdots, \bm{x}_I$, where $\bm{h}_k$ is $k$th row of basis matrix $\bm{H}$ and $\bm{x}_i$ is $i$th row of observed data matrix $\bm{X}$. Specifically, each basis vector $\bm{h}_k = \sum_{i = 1}^{I}v_{ki}\bm{x}_i$, where $\sum_iv_{ki} = 1, v_{ki} \geq 0$, $i = 1, \cdots, I$. In matrix notation, it can be expressed as  $\bm{H} = \bm{V}\bm{X}$. Furthermore, data points $\bm{x}_1, \cdots, \bm{x}_I$ are convex combination of basis vectors $\bm{h}_1, \cdots, \bm{h}_K$. Thus, AA  can be expressed as follows:
\begin{equation*}
\begin{split}
    & \bm{X} \approx \bm{U}\bm{VX}\\
    \text{subject to} ~~~ &\bm{U}\bm{1} = \bm{1},~~~  \bm{V}\bm{1} = \bm{1}
\end{split}    
\end{equation*} 
where $\bm{U}$ and $\bm{V}$ are nonnegative matrices of size $I \times K$ and $K \times J$, respectively. AA is general unique \citep{cutler1994archetypal, morup2012archetypal, javadi2020nonnegative, JMLR:v25:21-0233}.

When $\bm{V}$ includes a diagonal matrix with positive elements up to permutation, AA can be taken as NMF under the separability assumption. Furthermore, new objective functions that combine the advantages of both AA and NMF have been proposed \citep{De2020Near, javadi2020nonnegative, JMLR:v25:21-0233}. For example, \citet{javadi2020nonnegative} optimizes a trade-off between two objective functions: the distance of data points from the convex hull of the basis vectors (which can be seen as NMF data fitting error) and the distance of the basis vectors from the convex hull of the data points (which can be seen as a regularization).

\section{Conclusion}\label{S: Discussion and conclusion}

This paper has four goals. First, we illustrate that EMA from sedimentary geology, LBA from social science, asymmetric PLSA from machine learning have the same parametric form, and they are equivalent to symmetric PLSA from machine learning and LCA for two-way tables from social science and are a special case of NMF from machine learning. Second, we study an essential problem: the identifiability. We prove that the solution of NMF is unique if and only if the solution of LBA, LCA, EMA, and PLSA is unique. Existing results regarding uniqueness properties for NMF are worked out in much more detail than for LBA, LCA, EMA, and PLSA. Thus, these results for NMF are also useful for LBA, LCA, EMA, and PLSA. Third,  we provide a brief review of the algorithms of these models. Fourth, we analyse a sedimentary geology dataset using NMF. The data and code for this paper are on the GitHub website \url{https://github.com/qianqianqi28/NMF-and-related-methods-sedimentary-geology}.

\section*{Data available statement}

The Data and code are on the GitHub website \url{https://github.com/qianqianqi28/NMF-and-related-methods-sedimentary-geology}.

\section*{Acknowledgements}

We wish to thank Petra Goessen (Hoogheemraadschap Hollands Noorderkwartier HHNK) for initiating and funding the Slufter monitoring project, Erik van der Spek and Thomas van Es (Staatsbosbeheer) for support and providing access permits. Daan Eldering and Eke van Leeuwen (former Earth Science students, Vrije Universiteit Amsterdam) are thanked for executing the field work and grain-size analyses, Simon Troelstra, Jan-Berend Stuut and Martine Hagen (Vrije Universiteit Amsterdam) for co-supervision of the project. We gratefully acknowledge Nicolas Gillis (University of Mons) for his comments and suggestions that helped improve this paper.

\section*{Fundings}

Maarten A. Prins is supported by Hoogheemraadschap Hollands Noorderkwartier.

\section*{Disclosure statement}

No potential conflict of interest was reported by the authors.

\bibliography{reference.bib}

\end{document}